%% file: main.tex
\documentclass[journal]{IEEEtran}
\usepackage{graphicx}
\usepackage{amsmath,amssymb} 
\usepackage{color}
\usepackage{epsfig}
\usepackage{algpseudocode}
\usepackage{comment}
\usepackage[ruled,vlined]{algorithm2e}
\usepackage{booktabs}
\usepackage{tabu}
\usepackage{ntheorem}
\newtheorem{theorem}{Theorem}
\usepackage{subfig}

\newcommand{\vect}[1]{\mathbf{#1}}

\newcommand{\set}[1]{\mathcal{#1}}

\newcommand{\mybullet}[1]{\noindent \textbf{#1}}

\begin{document}
%
\title{Exploring Spatial Diversity for Region-based Active Learning}
%
%
%

\author{Lile Cai,
        Xun Xu,~\IEEEmembership{Senior Member,~IEEE,}
        Lining Zhang,~\IEEEmembership{Member,~IEEE,}
        Chuan-Sheng Foo
\thanks{L. Cai, X. Xu, L. Zhang and C.S. Foo are with the Institute for Infocomm Research (I2R), A-STAR, Singapore. e-mail: \{caill,xu\_xun,zhang\_lining,foo\_chuan\_sheng\}@i2r.a-star.edu.sg.}
}

%
%

\markboth{IEEE Transactions on Image Processing, 2021}%
{Shell \MakeLowercase{\textit{et al.}}: Bare Demo of IEEEtran.cls for IEEE Journals}
%



\maketitle

\begin{abstract}
State-of-the-art methods for semantic segmentation are based on deep neural networks trained on large-scale labeled datasets. Acquiring such datasets would incur large annotation costs, especially for dense pixel-level prediction tasks like semantic segmentation. We consider region-based active learning as a strategy to reduce annotation costs while maintaining high performance. In this setting, batches of informative image regions instead of entire images are selected for labeling. Importantly, we propose that enforcing local spatial diversity is beneficial for active learning in this case, and to incorporate spatial diversity along with the traditional active selection criterion, e.g., data sample uncertainty, in a unified optimization framework for region-based active learning. We apply this framework to the Cityscapes and PASCAL VOC datasets and demonstrate that the inclusion of spatial diversity effectively improves the performance of uncertainty-based and feature diversity-based active learning methods. Our framework achieves $95\%$ performance of fully supervised methods with only $5-9\%$ of the labeled pixels, outperforming all state-of-the-art 
region-based active learning methods for semantic segmentation.
\end{abstract}


%
\IEEEpeerreviewmaketitle

\section{Introduction}
\input{introduction.tex}

\section{Related Works}
\input{related_work.tex}

\section{Methodology}
\input{method.tex}

\section{Experiments}
\label{sec_exp}
\input{experiments.tex}

\section{Analysis of design choices and algorithm behaviour}
\input{discussions.tex}

\section{Conclusion}
\input{conclusions.tex}

\vspace{0.5em}
\section*{Acknowledgment} This research is supported by the Agency for Science, Technology and Research (A*STAR) under its AME Programmatic Funds (Grant No. A20H6b0151). The computational work for this article was partially performed on resources of the National Supercomputing Centre, Singapore (https://www.nscc.sg).


%

\appendices
\input{supplemental}



\ifCLASSOPTIONcaptionsoff
  \newpage
\fi



%
\bibliographystyle{IEEEtran}
\bibliography{references}

%
\vspace{-4em}
\begin{IEEEbiography}[{\includegraphics[width=1in,height=1.25in,keepaspectratio]{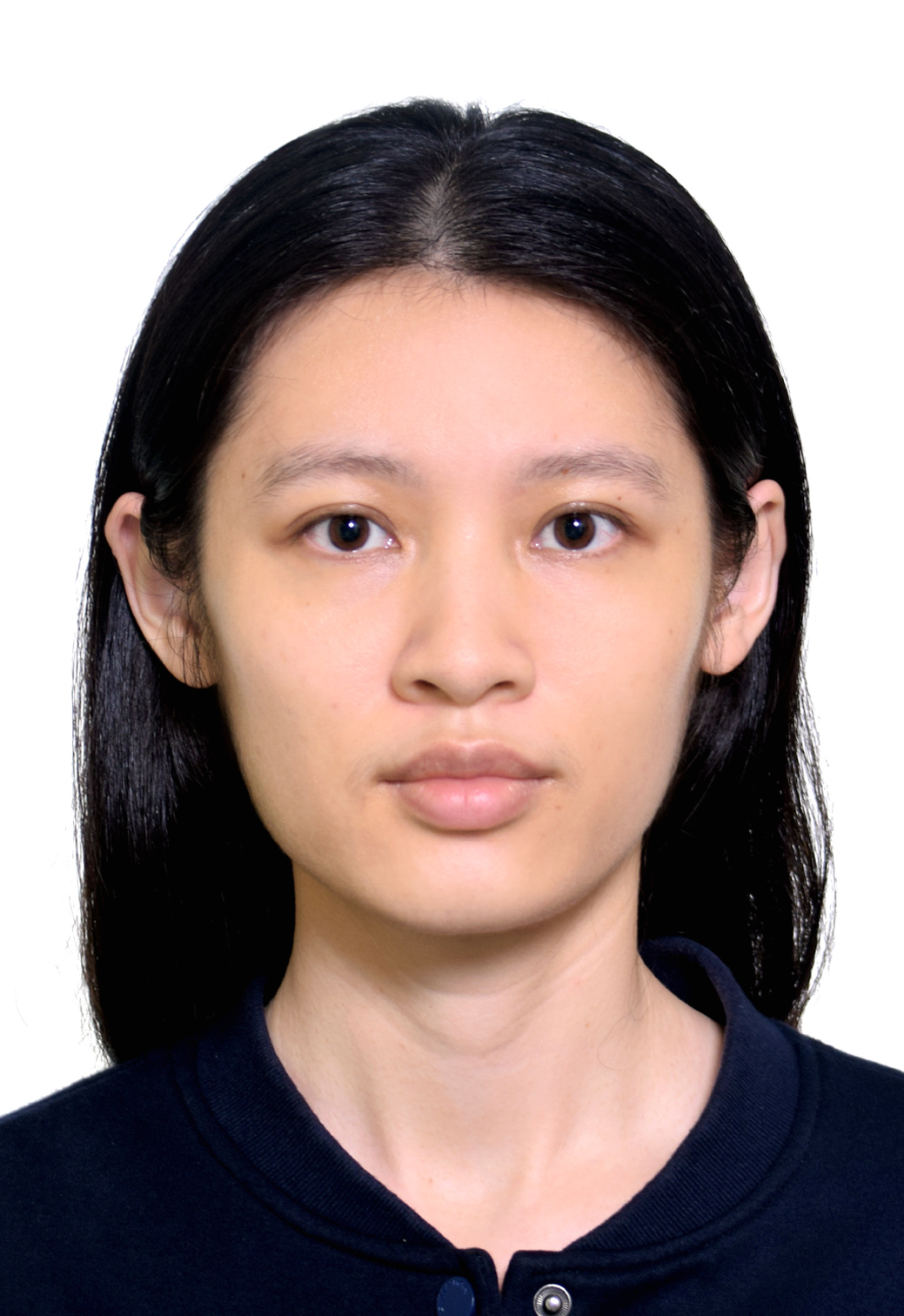}}]{Lile Cai}
received her Bachelor of Engineering from the University of Science and Technology of China in 2011 and Ph.D. from the National University of Singapore in 2015. She has been a research scientist with Institute of Infocomm Research since 2016. Her research interests include computer vision and data-efficient learning.
\end{IEEEbiography}

\vspace{-5em}
\begin{IEEEbiography}[{\includegraphics[width=1in,height=1.25in,keepaspectratio]{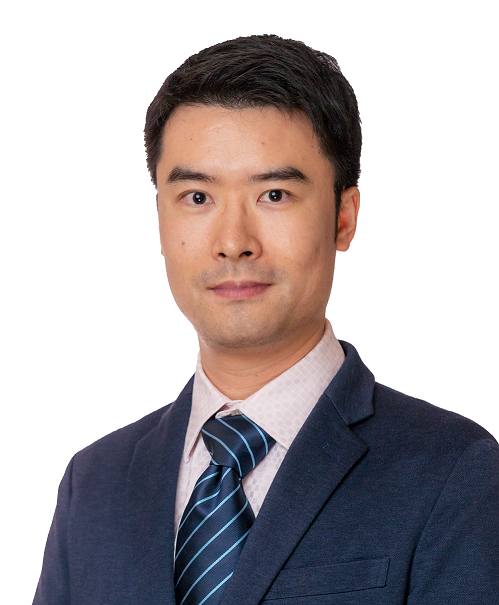}}]{Xun Xu}
received the PhD degree from Queen
Mary University of London in 2016. He is now with
Institute of Infocomm Research (I2R), A*STAR. He was a research fellow with National University
of Singapore from 2016 to 2019.  His research interests include semi-supervised learning,
active learning, adversarial learning, 3D point cloud,
motion segmentation, etc.
\end{IEEEbiography}

\vspace{-5em}
\begin{IEEEbiography}[{\includegraphics[width=1in,height=1.25in,clip,keepaspectratio]{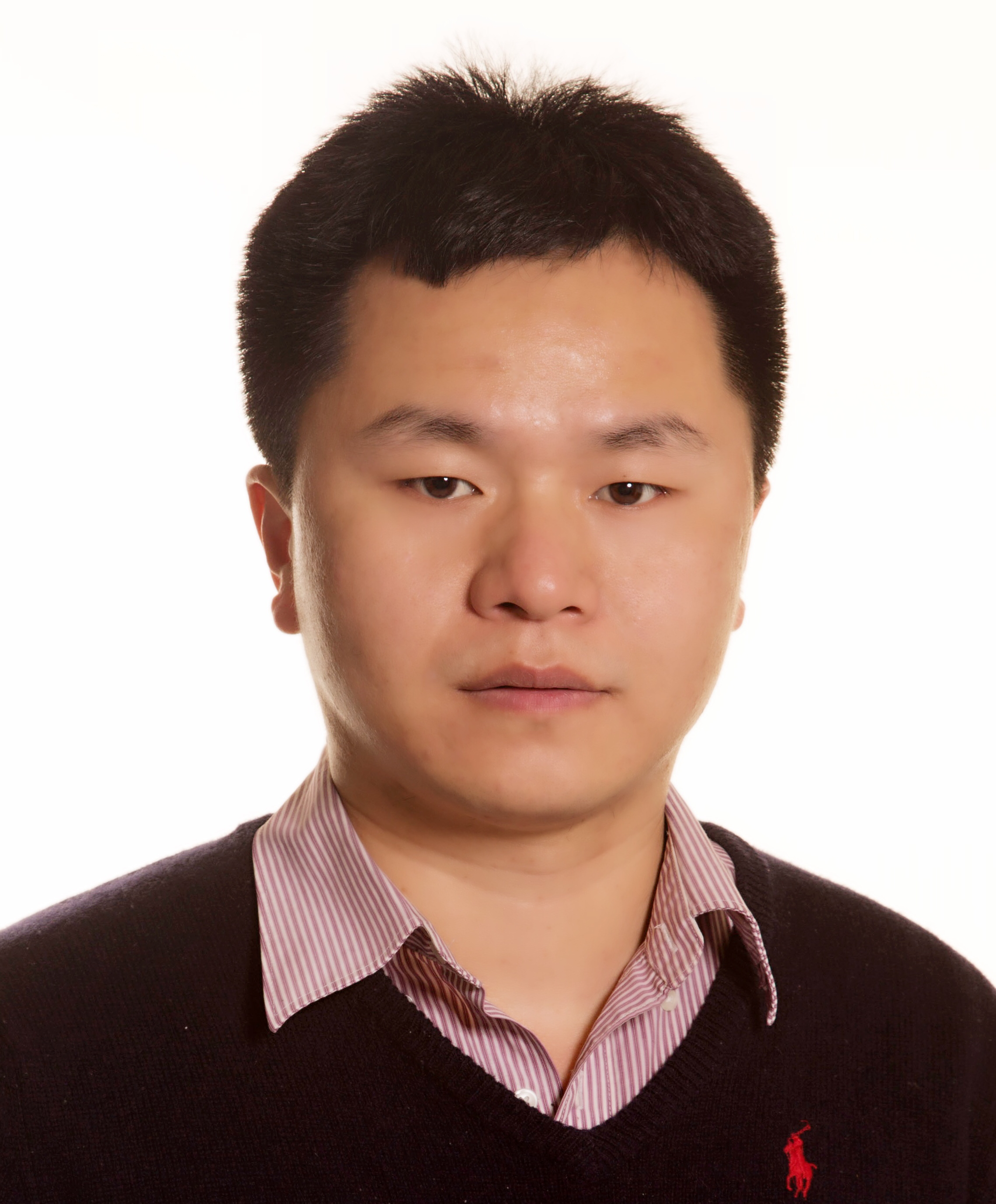}}]{Lining Zhang} is currently a research scientist with Institute for Infocomm Research, A*STAR, Singapore. He received his BEng, MSc and PhD degree in Electronical Engineering from the Xidian University, Xi’an, China and the Nanyang Technology University, Singapore, respectively. His research interests include multimedia information retrieval, video/image processing, learning with less data, etc. He is a member of IEEE.  
\end{IEEEbiography}

\vspace{-5em}
\begin{IEEEbiography}[{\includegraphics[width=1in,height=1.25in,clip,keepaspectratio]{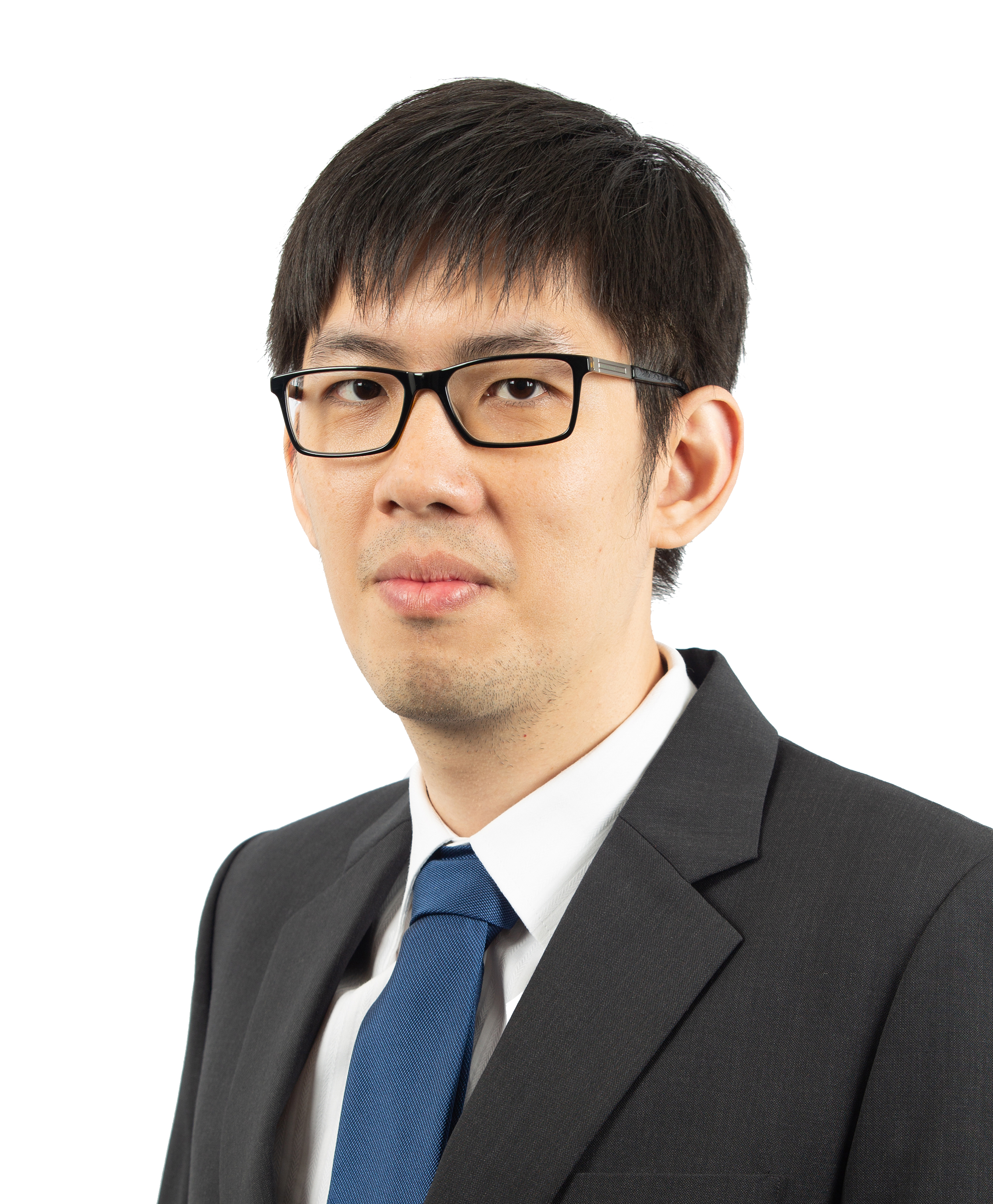}}]{Chuan-Sheng Foo} received his BS, MS and PhD degrees from Stanford University. He currently leads a research group at the Institute for Infocomm Research, A*STAR, Singapore, that focuses on developing data-efficient deep learning algorithms that can learn from less labeled data. 
\end{IEEEbiography}









\end{document}

%% file: introduction.tex
Deep convolutional neural networks (CNNs) trained on large collections of labeled images have achieved state-of-the-art performance on many computer vision tasks. However, assembling such large labeled datasets is time-consuming and expensive, especially so for more complex tasks such as semantic segmentation, where the semantic boundaries have to be precisely delineated by polygons to further produce pixel-wise labels.

Active learning (AL) offers a potential strategy for reducing this labeling burden, by selecting only a subset of the dataset to be labeled \cite{settles.book12,YangMNCH15,gal2017deep,SenerS18}. This subset should be both \emph{informative} and \emph{diverse} -- where informativeness is commonly characterized by high model uncertainty on the sample, while diversity is evaluated using distances between selected samples. In particular, diversity is essential to avoid overlap between similar samples, especially when batches of samples are labeled at a time. 

While active learning is a well-studied problem with a rich literature for standard classification tasks \cite{YangMNCH15,gal2017deep,SenerS18,settles.book12}, active learning for semantic segmentation is far less studied, especially in the context of CNNs. Semantic segmentation tasks have a number of unique characteristics that make straightforward adaptation of active learning approaches for standard classification tasks a sub-optimal approach. For instance, while labels are defined at the pixel-level, treating each pixel as a sample in an AL framework will dramatically increase annotation costs as compared to polygon-based labeling schemes typically used in practice \cite{Cordts2016Cityscapes}. On the other hand, per-image annotation is sub-optimal as large areas within an image typically contain identical labels. As such, in order to further optimize the use of labeling resources, recent works have considered the selection of regions instead of entire images or individual pixels for active learning \cite{MackowiakLGDLR18,Casanova2020}.

\begin{figure}[t]
    \centering
    \includegraphics[width=\linewidth]{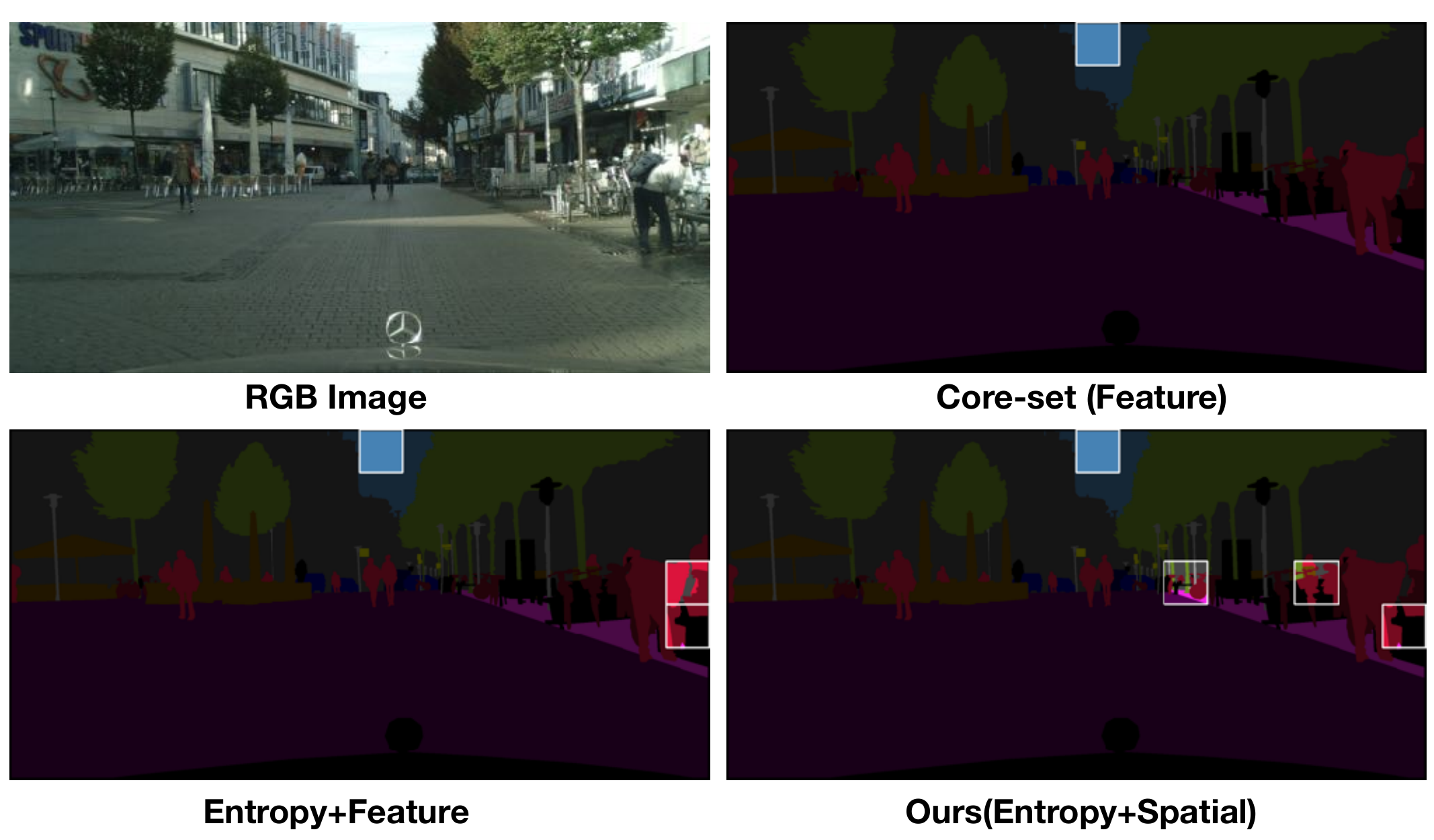}
    \caption{We propose a new objective, i.e., spatial diversity, for the selection of unlabeled samples in region-based AL. Traditional feature diversity (Core-Set) and uncertainty + feature diversity (Entropy+Feature) ignore spatial cues that are important for selecting samples covering more categories.}
    \label{fig:Intro}
 
\end{figure}

In this work, we focus on this region-based active learning setting for semantic segmentation, and propose that enforcing \emph{local spatial diversity} is beneficial in this case. Our proposal is motivated by several observations: 1) Spatially adjacent regions typically have similar labels and share similar features. Information diversity would be compromised when nearby regions are simultaneously selected for labeling (Fig.~\ref{fig:Intro}); 2) Spatial diversity, defined only by a region's coordinates, does not require a good feature extractor in contrast to feature-based diversity measures, and is able to help with inaccurate uncertainty estimates at the early iterations of active learning when only a small number of regions are labeled; 3) Neurons at deeper layers of CNNs have a large receptive field, and neighboring regions typically have similar context, while regions far away from each other or located in different images typically have different context and thus can be beneficial for CNN training even if they have similar features. Technically, we introduce a piece-wise constant distance function to strongly penalize close-by region pairs while not explicitly encouraging faraway region pairs. Our proposed spatial diversity measure is compatible with and complements existing uncertainty and feature-space diversity measures. We combine these measures in a unified optimization framework for selecting a batch of regions to be labeled using an efficient greedy algorithm that scales to the large ($\sim 10^5$) number of regions in benchmark semantic segmentation datasets.

We summarize the contributions of this work as follows:
\begin{itemize}
    \item We propose spatial diversity as a new objective for region-based AL, which can replace the traditional feature diversity objective as a more effective strategy to enforce diversity in selected samples. To the best of our knowledge, this is the first explicit use of spatial diversity for region-based AL.
    \item We formulate a unified optimization framework for jointly optimizing model uncertainty and diversity in an efficient way using a greedy algorithm that scales to region counts ($\sim 10^5$) in segmentation datasets.
    \item Our proposed method Entropy+Spatial achieves the state-of-the-art performance on the Cityscapes and PASCAL VOC~2012 datasets, achieving $95\%$ performance of fully supervised methods with only $8.4\%$ and $5.9\%$ pixels labeled, respectively.
\end{itemize}

%% file: related_work.tex
\noindent\textbf{Classical AL}:
As an established approach to mitigate the problem of limited data annotation, AL has been extensively studied \cite{settles.book12}. According to the criteria used to select data samples for labeling, various approaches can be classified into three categories: uncertainty-based, diversity-based and expected change (EC)-based schemes. Uncertainty-based approaches select data samples that the current model is most uncertain of, where uncertainty can be measured by the posterior probability of a binary classifier \cite{LewisG94}, the entropy of class posterior probabilities \cite{JoshiPP09}, best-versus-second best margin \cite{JoshiPP09}, and the disagreement among several models \cite{houlsby2011bayesian}. Diversity-based approaches take into account the distribution of all unlabeled data to make the selected data as diverse as possible. Yang et al. \cite{YangMNCH15} formulated the uncertainty sampling as an optimization problem and added diversity constraint as a regularizer. Nguyen and Smeulders \cite{NguyenS04} performed clustering on the dataset and the selection criterion was biased towards the samples that are not only close to the classification boundary but also cluster representatives. Wang and Ye \cite{WangY13} applied the empirical risk minimization principle to active learning and by minimizing the proposed upper bound for the true risk, their method was able to select the most informative samples while preserving the source distribution. EC-based approaches aim to select those samples that once their labels are known, the expected change in the model prediction is largest. Vezhnevets et al. \cite{VezhnevetsBF12} built a conditional random field (CRF) on superpixels and selected nodes that can induce the largest EC of the CRF output for labeling.

\noindent\textbf{AL for Deep CNNs}:
Deep CNNs have successfully addressed a variety of computer vision tasks, and it is known that CNNs require large amount of labeled data to train. As a result, active learning has been specifically investigated in the deep learning context to alleviate data annotation efforts. Among these works, Beluch et al. \cite{BeluchGNK18} trained an ensemble of classifiers to estimate the uncertainty of unlabeled data. Yoo and Kweon \cite{Yoo2019} proposed to use loss as an indication of uncertainty. The core-set approach \cite{SenerS18} did not use any uncertainty information and instead relied on data distribution for active selection. Zhou et al. \cite{ZhouSZGGL17} combined uncertainty and diversity in their selection criterion and computed the two metrics on the augmentation data of a candidate image. These methods are targeted for image classification tasks and do not consider the challenges uniquely presented in region-level active learning, e.g., spatial relationship of regions and large region number, and thus a naive application to semantic segmentation task would be sub-optimal.

\noindent\textbf{AL for CNN-based Semantic Segmentation}:
Only a handful of works have investigated AL for image segmentation \cite{dutt2016active,YangZCZC17,mahapatra2018efficient,sinha2019variational,Casanova2020,kasarla2019region,siddiqui2020viewal}. These works can be classified into image-level methods and region-level methods, based on the granularity of data annotation. Image-level methods select the entire image for annotation, while region-level methods divide an image into patches and select specific patches for annotation. Among the first category, Yang et al. \cite{YangZCZC17} considered AL for biomedical image segmentation, which is often a simpler foreground/background segmentation task in comparison with the more complex semantic segmentation task we deal with in this work. It measured the uncertainty of an image by deep CNN models, and compute similarity between two images using CNN based image descriptors. Active selection was carried out iteratively by selecting the images that maximize the representativeness of the selected set. Sinha et al.\cite{sinha2019variational} trained a variational auto-encoder (VAE) in an adversarial manner to learn a latent space to discriminate between labeled/unlabeled samples and employed the discriminator's prediction score to select unlabeled samples. Among the later category, CEREALS \cite{MackowiakLGDLR18} investigated how different methods (Random, Entropy and Vote Entropy) behaved when the cost was measured in pixels vs. clicks and proposed to consider annotation cost in the selection procedure. However, clicked data was needed to train the cost model, which is not usually available for a public dataset. Kasarla et al. \cite{kasarla2019region} performed active learning on irregularly-shaped regions, i.e., superpixels, and entropy was employed as the selection strategy. Spatial diversity has not been explored for active selection in \cite{kasarla2019region}. ViewAL \cite{siddiqui2020viewal} exploited the prediction consistency across different views to select informative regions for semantic segmentation in multi-view datasets. Recently, Casanova et al. \cite{Casanova2020} employed reinforcement learning to learn a region selection policy for image segmentation. However, additional labeling data was required to train the policy network which significantly increased the labeling budget. In this work, we stick to the conventional approaches by introducing the spatial diversity objective into existing selection measures. It inherits the advantages of being explicit and does not require additional training data.

%% file: method.tex
We first present a unified optimization framework for region-based AL that encompasses several existing AL approaches and an efficient greedy algorithm to solve it. We then introduce our spatial diversity objective and elaborate the rationale of specific choices.

\subsection{Problem Formulation and Greedy Solution}\label{sect:uni_obj}
Active learning is an iterative process where a batch of samples to be labeled is selected at each iteration; the process is repeated until the labeling budget is exhausted. At iteration $t$, we denote the already labeled set of samples as $\set{L}_t$ and unlabeled set as $\set{U}_t$. The goal is then to select a batch of $K$ informative samples, $\set{B}_t\subseteq\set{U}_t$, where an informative batch is characterized as including a \emph{diverse} set of samples that the model is \emph{uncertain} about. We formulate this as an optimization problem, where the following objective is maximized,

\begin{equation}\label{eq:uni_obj}
    \max_{\set{B}_t\subseteq\set{U}_t} \lambda_u \Omega(\set{B}_t) + \Phi(\set{L}_t,  \set{B}_t,  \set{U}_t), \quad s.t.\quad |\set{B}_t| = K.
\end{equation}
Here, $\Omega(\cdot)$ and $\Phi(\cdot)$ denote set functions respectively defined on uncertainty measurements $u(x)$ of sample $x$ and diversity measurements $d(x,y)$ between two samples $x$ and $y$. Such a unified objective can be instantiated as various existing AL approaches. In particular, the Core-Set method \cite{SenerS18} is a special case of Eq.~(\ref{eq:uni_obj}) with $\lambda_u = 0$ and $\Phi:-\max\limits_{\vect{x}_i\in\set{U}_t}\min\limits_{\vect{x}_j\in\set{L}_t\cup \set{B}_t} d(\vect{x}_i, \vect{x}_j)$. A variant related to the USDM method \cite{YangMNCH15} can be obtained with $\lambda_u=1$, $\Omega:\sum\limits_{\vect{x}_i\in\set{B}_t}u(\vect{x}_i)$ and $\Phi:\sum\limits_{\vect{x}_i, \vect{x}_j\in \set{L}_t\cup \set{B}_t}d(\vect{x}_i, \vect{x}_j)$. We see that the Core-Set method ignores uncertainty measurements, which may result in the selection of uninformative samples. On the other hand the original USDM method, while simultaneously accounting for uncertainty and feature diversity, formulated an optimization problem in a way that is difficult to solve efficiently for large datasets. In contrast to both methods, we are inspired by the methods developed for information retrieval which encourages retrieved documents to be simultaneously relevant and diverse \cite{ravi1994heuristic,gollapudi2009axiomatic} and propose to adopt the following two set functions,
\begin{equation}
    \Omega:\min\limits_{\vect{x}_i\in\set{B}_t }u(\vect{x}_i),\quad \Phi:\min\limits_{\vect{x}_i, \vect{x}_j \in\set{L}_t\cup\set{B}_t}d(\vect{x}_i, \vect{x}_j).
\end{equation}

The rationale behind our choice is threefold. Firstly, the unary term involves maximizing the minimum uncertainty of selected regions. This is a stronger constraint compared to the sum of uncertainties used in USDM \cite{YangMNCH15} since our objective assumes those samples with low entropy would not be selected even if they are far away from selected samples in feature or spatial distance. Secondly, the pairwise term involves maximizing the minimum distance between selected samples. This encourages visually similar regions or local regions to not be selected simultaneously thus ensuring diversity. Finally, we can reformulate the overall objective as a single \textit{Max-Min} problem as in Eq.~(\ref{eq:obj_maxmin}), which can be efficiently solved (approximately) by a greedy algorithm \cite{gollapudi2009axiomatic}.

\begin{equation}\label{eq:obj_maxmin}
   \max_{\set{B}_t\subseteq\set{U}_t}\min_{\substack{\vect{x}_k \in \set{B}_t,\\ 
   \vect{x}_i, \vect{x}_j \in\set{L}_t\cup\set{B}_t}} 
   \left[\lambda_u u(\vect{x}_k) + d(\vect{x}_i,\vect{x}_j) \right].
\end{equation}

We use entropy for the uncertainty measure, i.e., $u(\vect{x}_i)=\sum_j H(x_{ij})$, where $H(x_{ij}) = -\sum_c p({x}_{ij}^c)\log p({x}_{ij}^c)$ with $x_{ij}$ denoting the $j$-th pixel of the $i$-th region, $c$ denoting the class and $p(\cdot)$ denoting the per-pixel class posterior from the segmentation network. The diversity measure $d(\vect{x}_i,\vect{x}_j)$ can be feature diversity, spatial diversity or a combination of these; we discuss specific choices in Section~\ref{sect:Method_SpatDist} and Section~\ref{sect:comp_methods}.
The greedy algorithm begins with a randomly initialized labeled set $\set{L}_0$ and proceeds for $T$ iterations. During each iteration, we select a sample $\vect{x}$ from the candidate pool that maximizes a potential function $\phi(\vect{x})$ as defined in Eq.~(\ref{eq:pot_func}) and repeat the selection process until we select the desired number of samples for a batch. After each iteration, the new batch of labeled samples is combined with all existing labeled samples and used to retrain the segmentation network. This procedure is summarized in Algorithm~\ref{greedy_algo}. 

\begin{equation}\label{eq:pot_func}
\begin{split}
        \phi(\vect{x})&=\min_{\vect{x}_i\in\set{L}_t\cup\set{B}_t} \lambda_u u(\vect{x}) +  d(\vect{x}_i,\vect{x})\\
        &=\lambda_u u(\vect{x}) + \min_{\vect{x}_i\in\set{L}_t\cup\set{B}_t}  d(\vect{x}_i,\vect{x}).
\end{split}
\end{equation}
 
\begin{algorithm}
\SetAlgoLined
\SetKwInOut{Input}{Input}
\SetKwInOut{Output}{Output}
\Input{Initial labeled set of regions $\set{L}_0$, Initial unlabelled set of regions $\set{U}_0$, Batch-size $K$, Maximum number of batches $T$}
\Output{Output labeled set of regions $\set{L}_T$}
$t=0$\;
\While{$t \leq T$}
{
    Train segmentation network $f(\vect{x})$ on $\set{L}_t$ until convergence\;
    $\set{B}_t=\emptyset$\;
    \While{$|\set{B}_t|<K$}
    {
        $\hat{\vect{x}}=\arg\max\limits_{\vect{x}\in\set{U}_t} \phi(\vect{x})$\;
        $\set{B}_t=\set{B}_t\cup \hat{\vect{x}}$\;
        $\set{U}_t=\set{U}_t \setminus \hat{\vect{x}}$\;
       
    }
    $\set{L}_t=\set{L}_t\cup \set{B}_t$\;
    $t = t + 1$\;
    
}
\caption{Greedy Active Selection}\label{greedy_algo}

\end{algorithm}

\subsection{Spatial Diversity Objective}\label{sect:Method_SpatDist}
The aforementioned active selection objective considers balancing selecting regions of low model confidence and maximizing diversity. To the best of our knowledge, all previous works adopt a distance defined in feature space for diversity maximization \cite{SenerS18,YangZCZC17,YangMNCH15}, which is typically defined as the $L_2$ distance between feature vectors $f(\vect{x})$: $d_f(\vect{x}_i,\vect{x}_j)=||f(\vect{x}_i)-f(\vect{x}_j)||_2$. However, this feature diversity measure may not be reliable because the feature extractor is either trained with very few labeled samples, thus being non-stable, or uses a network pre-trained on another dataset, which may not be ideal due to domain shift. Moreover, previous works \cite{sinha2019variational,ren2020survey} have found that feature diversity did not perform well on high dimensional data due to the distance concentration phenomenon. In this section, we introduce a spatial diversity objective to replace the feature diversity objective. It aims to enforce selected regions to be diverse in the image domain therefore there is a high chance of covering diverse categories.

A straightforward way to define spatial distance is as the $L_p$ norm between the coordinates $\vect{loc}$ of the centers of the two regions,
\begin{equation}\label{eq:linear_dist}
    d_s(\vect{x}_i,\vect{x}_j)=||\vect{loc}_i - \vect{loc}_j||_p.
\end{equation}
With such a distance, by optimizing Eq.~(\ref{eq:obj_maxmin}) we are essentially encouraging selecting regions as far away as possible. However, this may not be desirable as it may end up selecting far-away but low-entropy regions. The key is to remove redundancy in a local neighborhood. To remedy this issue, we propose to use the piece-wise distance function as defined in Eq.~(\ref{eq:step_dist}), where $a, b, c$ and $\tau$ are predefined constants. When setting $a, b, c$ we follow two principles : 1) $a<b$, so that candidate regions close enough (less than $\tau$) to a selected region are less favored than other candidates at least $\tau$ faraway from any selected region. In contrast, if some regions are at least $\tau$ faraway from any selected regions, there should be no preference between them; 2) $c>=b$, as regions from different images should have a distance larger than or equal to the distance if they are from the same image. In our experiment, we set $a=1, b=2, c=2$ and $\tau=N$ where $N$ is region size and choose $L_\infty$ norm for the distance calculation. 

\begin{equation}\label{eq:step_dist}
\small
    d_s(\vect{x}_i,\vect{x}_j)=\begin{cases}
        0 & \text{if } i=j, \\
        a & 
        ||\vect{loc}_i-\vect{loc}_j||_p \leq \tau  \text{ \& $i$, $j$ from the same image,}\\
        b & ||\vect{loc}_i-\vect{loc}_j||_p > \tau \text{ \& $i$, $j$ from the same image,}\\
       c & \text{$i$, $j$ from different images.}
    \end{cases}
\end{equation}

Note that the spatial diversity term introduced here will not eliminate the selection of candidate regions with distance less than $\tau$ to a selected region and thus is not equivalent to a checkerboard selection pattern. This is because each term in the objective will be linearly normalized to [0,1] before summation and all the terms (i.e., entropy, feature diversity and spatial diversity) will play a role. If a region has very high entropy, e.g., close to 1, it can still get selected even if it is a neighbor of a selected region (i.e., spatial distance is 0.5).

%% file: experiments.tex
In this section, we first introduce the datasets we experimented with, the implementation details and evaluation metric. Then we show comparisons with the state-of-the-art methods and our own variants. Finally, we present some qualitative results and give additional insights.

\subsection{Datasets}
\noindent\textbf{Cityscapes} \cite{Cordts2016Cityscapes} was developed for urban environment scene understanding. It contains 19 categories and the number of images for the train/validation/test split is 2975/500/1525. The image dimension is $1024\times 2048$. Images from this dataset are captured by a vehicle traversing the cities of Europe and are typically complex street scenes. We conduct active learning on the training set and report the performance on the validation set.

\noindent\textbf{PASCAL VOC~2012} \cite{pascal-voc-2012} has been widely used for evaluating image segmentation tasks. The dataset consists of 1464 images for training and 1449 images for validation. The maximum image dimension is 500. This dataset contains 20 object categories and each image typically contains only several dominant objects  like dog, cat and bottle. We conduct active learning on the training set and report the performance on the validation set.

\subsection{Implementation Details}\label{sect:imp}

\noindent\textbf{Fully supervised training details}:
We first train a segmentation model using the fully labeled training set, which serves as a performance upper bound for all AL methods. We follow the same training protocol in auto-deeplab \cite{liu2019auto}: the base learning rate is decayed by the ``poly" policy (where the learning rate is multiplied by $(1 - \frac{iter}{max_iter})^{power}$ with $power=0.9$). We set the base learning rate as 0.01 for Cityscapes and 0.007 for VOC~2012 and . We train the model for 30000 iterations with a batch size of 16. For data augmentation, the image is first randomly rescaled by a factor between 0.5 and 2 and then randomly cropped to ($513\times513$). During testing, the original size image is fed into the model and the prediction results of this single-scale inference are used to compute mean Intersection Over Union (mIoU). 

\noindent\textbf{Segmentation model}: We used the open-source DeepLabv3+ \cite{deeplabv3plus2018} with the Xception-41 \cite{Chollet17} backbone as our segmentation model. The Xception-41 backbone network has been pre-trained on ImageNet \cite{deng2009imagenet}, but the segmentation model has not been pre-trained on the target segmentation datasets, i.e., all the AL methods start from the ImageNet pre-trained model. DeepLabv3+ employs an encoder-decoder structure with atrous convolution in encoder for multi-scale feature extraction. We use the default settings for DeepLabv3+: the output stride of the prediction is set to 16, which is refined by the decoder to a output stride of 4; the rates of atrous convolution in the atrous spatial pyramid pooling are 6, 12 and 18.

\noindent\textbf{Region division scheme}: Each image is divided into non-overlapping regions with a region size of $N\times N$ pixels. We choose $N=128$ for Cityscapes and $N=32$ for PASCAL VOC~2012. 

\noindent\textbf{Region feature extraction}: To enforce feature space diversity, each region is represented by a high-dimensional feature vector \cite{Casanova2020}. The feature can either be extracted from the current segmentation model or an independent model pre-trained on other tasks (e.g., image classification models trained on ImageNet). The latter may not be ideal due to the domain shift between ImageNet and the segmentation datasets, PASCAL VOC~2012 and Cityscapes. In our experiment, we use the last layer before prediction of the segmentation model as the feature extractor (i.e., decoder features, which are a combination of lower layer and higher layer 
activations). Features are extracted in every AL iteration using the model trained in the previous iteration. For each region, the feature values within that region are averaged to form the feature vector. To reduce computational cost and remove redundancy, PCA is applied to the feature vectors to reduce the dimension from 256 to 128.

\noindent\textbf{Batch training details}: We experiment with 6 batches (iterations) indexed from 0. For batch 0, we randomly select $1000$ regions and all comparing methods share the same model trained on the same initial batch. For batch $t=1,\ldots, 5$, $(2^t-2^{(t-1)})\times 1000$ regions are selected. After batch selection and labeling, the segmentation model is trained from scratch using all the data labeled so far (i.e., $2^t\times 1000$ regions). The model is trained for 100 epochs and other training details are the same as those in fully supervised training.

\subsection{Competing Methods}
\label{sect:comp_methods}
For brevity, we define $d(x_i, x_j)$ in Eq.~(\ref{eq:obj_maxmin}) as $\lambda_f d_f(\vect{x}_i,\vect{x}_j) + \lambda_s d_s(\vect{x}_i,\vect{x}_j)$, and control $\lambda_u$, $\lambda_f$ and $\lambda_s$ to create methods with different objectives.

\mybullet{Random:} In each training batch,  regions are randomly selected and incorporated into the labeled set $\set{L}_t$.

\mybullet{Entropy} ($\lambda_u=1, \lambda_f=0, \lambda_s=0$): It purely aims to select those regions with least confidence predicted by the network. The entropy of a region is computed as the average entropy of all pixels within that region.

\mybullet{Entropy+Random}: This strategy selects 50\% data by Entropy and the rest randomly sampled for each batch.

\mybullet{Feature} ($\lambda_u=0, \lambda_f=1, \lambda_s=0$ ): It aims to select image regions by maximizing the diversity in feature-space alone while disregarding entropy. Note that this variant is equivalent to the k-Center greedy solution for Core-Set \cite{SenerS18} and we thus denote it as \textbf{Core-Set}.

\mybullet{Feature+Spatial} ($\lambda_u=0, \lambda_f=1, \lambda_s=1$ ): It is a diversity-based method where both feature diversity and spatial diversity are taken into consideration.

\mybullet{Entropy+Feature} ($\lambda_u=1, \lambda_f=1, \lambda_s=0$): It aims to select regions by balancing entropy and feature-space diversity.

\mybullet{Entropy+Spatial} ($\lambda_u=1, \lambda_f=0, \lambda_s=1$): As opposed to the previous variant, this model aims to maximize spatial diversity instead of feature-space diversity.

\mybullet{Entropy+Spatial+Feature} ($\lambda_u=1, \lambda_f=1, \lambda_s=1$): It combines all three terms together and optimizes them jointly.

\subsection{Quantitative Results}
\noindent\textbf{Cityscapes}:
We present the active learning results on Cityscapes in Fig.~\ref{fig:al_curve}. We make the following observations from the results.
\begin{itemize}
    \item Entropy significantly outperforms Random. This implies that Random is not able to achieve good label diversity on this dataset. This is supported by the fact that this dataset is highly class-imbalanced: categories like rider, motorcycle, traffic light and bicycle are rare, or occupy a tiny fraction of images, while other categories like road, sky, building are common, or occupy a large spatial extent. Random selection will sample the most pixels from categories with large spatial extents and categories with small and rare objects are under-represented. Entropy can remedy this by selecting samples that the network is most uncertain of (which may mainly come from those small and rare objects).
    \item Feature diversity based methods (Core-Set, Entropy+Feature and Feature+Spatial) cannot beat the Entropy baseline, and Core-set which only considers feature diversity performs barely better than Random, implying that feature diversity cannot effectively solve the class imbalance problem. Feature+Spatial performs slightly better than Core-set, suggesting that combining feature and spatial diversity can improve the performance, but without considering the uncertainty information, it still cannot select sufficiently good samples.
    \item Spatial diversity can further boost the performance of Entropy, as seen when Entropy+Spatial consistently outperform Entropy after 4k budget. This indicates that there is still some redundancy in the samples selected by Entropy. Spatial diversity can help to remove this redundancy and the saved budget can then be used to select regions with a different context, facilitating pixel and label diversity.
    \item Entropy+Random performs somewhere between Random and Entropy. Overall we found that the better performing method (i.e., Random or Entropy depending on the datasets) is the upper bound for this mixture sampling scheme. These results suggest that Random is not able to provide the kind of diversity that Entropy+Spatial does. Entropy+Spatial allows the selected samples to still focus on high entropy regions while discouraging the selection of close-by regions. In comparison, Random will sample regions uniformly in space and most regions will be from classes occupying large spatial extents in the images like roads and buildings, which may not be as informative. 
    \item  Entropy+Spatial and Entropy+Feature+Spatial perform comparably: at a labeling budget of $8.4\%$ pixels, both can achieve more than $96\%$ accuracy of the fully supervised model. 
\end{itemize}

\begin{figure*}[htb]
    \centering
    
    \subfloat{\includegraphics[width=0.4\textwidth]{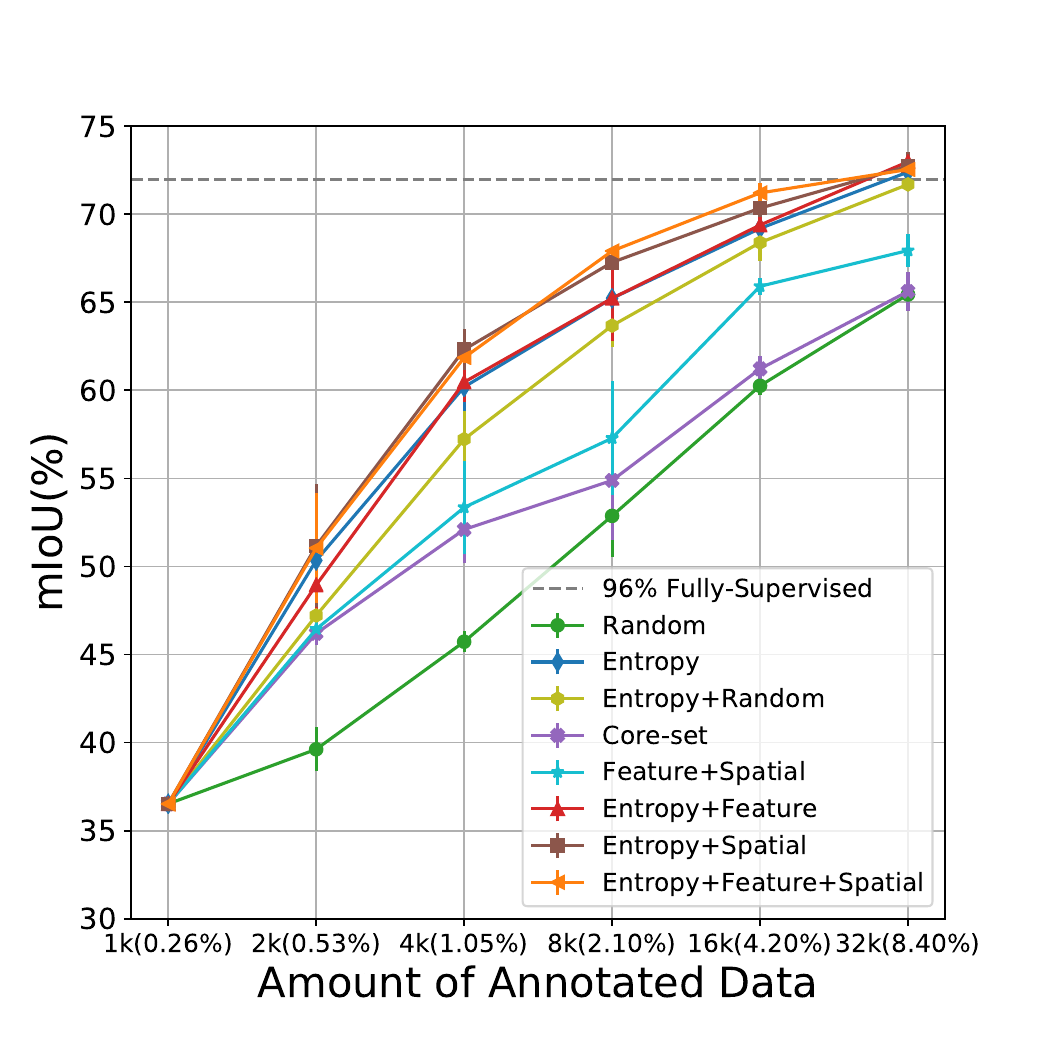}}
    \subfloat{\includegraphics[width=0.4\textwidth]{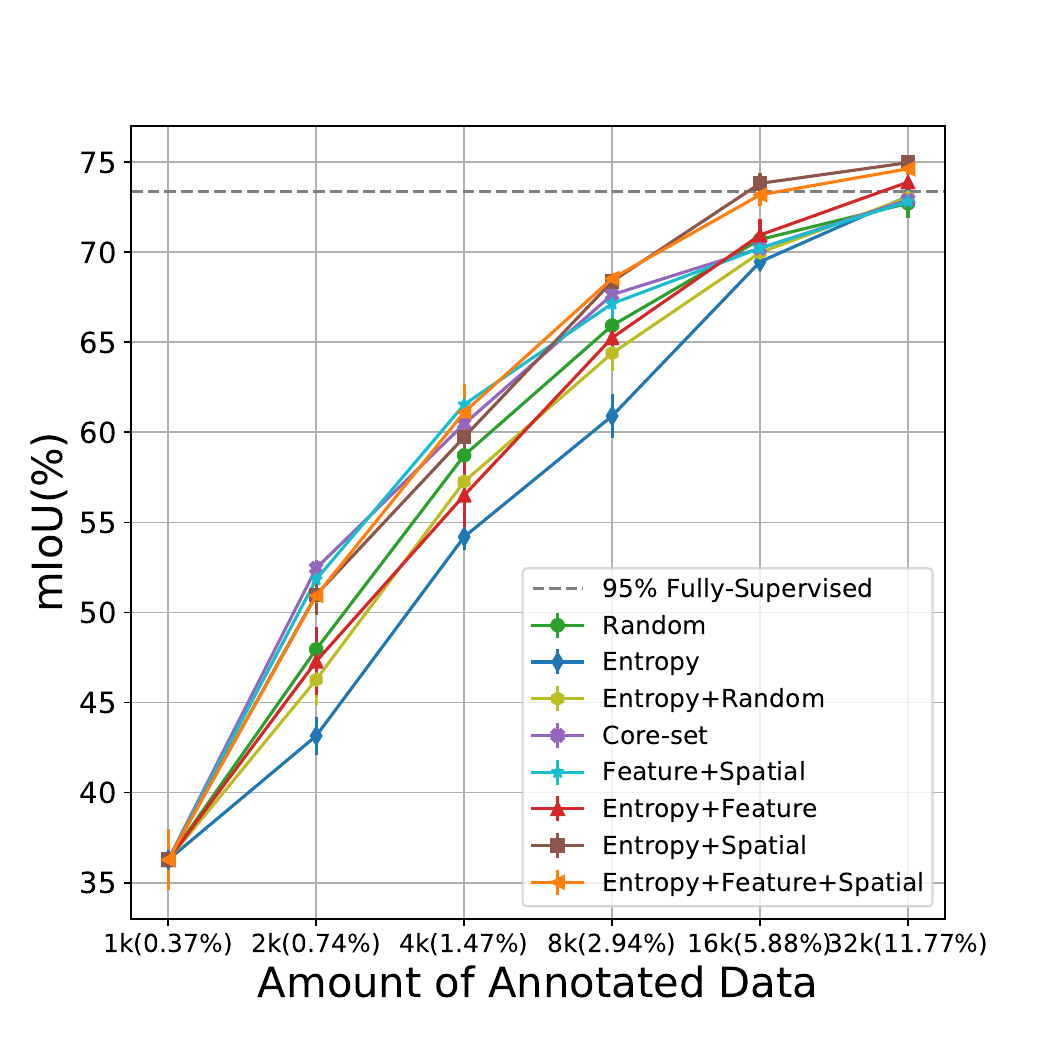}}
	\label{fig:al_curve}
    
	\caption{Segmentation performance vs. amount of annotated data on Cityscapes (left) and PASCAL VOC~2012 (right). We report the mean and standard deviation of 3 runs.}
    \label{fig:al_curve}
    \vspace{-1em}
\end{figure*}

\noindent\textbf{PASCAL VOC~2012}:
We evaluate on PASCAL VOC~2012 dataset in a similar way to Cityscapes. The results are presented in Fig.~\ref{fig:al_curve}. We make the following observations from the results:
\begin{itemize}
\item The Random baseline works reasonably well and even outperforms the Entropy method. As images in PASCAL VOC~2012 typically contain only several dominant objects, random sampling can achieve fairly good label and pixel diversity, while Entropy can end up selecting many neighboring regions as their uncertainty are similarly high and yields sub-optimal performance.
\item Feature diversity-based methods (Core-Set, Entropy+Feature and Feature+Spatial) cannot consistently beat Random, suggesting that these methods may select diverse but unimportant samples or even outliers for training.
\item Entropy+Random performs somewhere between Random and Entropy.
\item Entropy+Spatial and Entropy+Feature+Spatial consistently outperform the Random baseline and their counterparts that do not consider spatial diversity. With only $5.88\%$ of the pixels being labeled, both are able to achieve $95\%$ accuracy of the fully supervised model.
\end{itemize} 

\noindent\textbf{Comparison with a fixed budget}:
We further compare against the reinforced active learning (RAL) method for image segmentation \cite{Casanova2020} and a method based on Bayesian convolutional neural networks, i.e., Bayesian active learning by disagreement (BALD) \cite{gal2017deep} at a fixed budget. We adopt Entropy+Spatial as our final method as it performs comparably with Entropy+Feature+Spatial and has the benefits of not requiring feature engineering. For fairness, all the methods are run under the same backbone networks as used in \cite{Casanova2020} (which is an adaptation of feature pyramid network\cite{lin2017feature}) and the region size is $128\times 128$ for all methods. For RAL, the mIoU number reported in Table 1 of \cite{Casanova2020} where the budget is 12k regions is compared with ours. We note that for RAL, additional 350 labeled images are needed to train the policy network for reinforcement learning. These 350 labeled image are also used to pre-train the segmentation model and perform validation during the AL cycles. Therefore, the actual labeled budget is 56800 regions (12000 + 350*128 = 56800, 14.9\% of total pixels). All other methods can start from 0 labeled images and a model pre-trained on the synthetic GTA dataset \cite{richter2016playing} (no human labeling efforts needed). The actual annotation budget will be just the number of regions selected by AL. We follow our selection scheme, i.e., at batch 0, 1000 regions are selected, and at batch $t, t=1,\ldots, 5$, $(2^t-2^{(t-1)})\times 1000$ regions are selected. We report the results at batch 4, which is equivalent to a budget of 16k regions (4.2\% of total pixels). The results are summarized in Table~\ref{tab:fixedbudget}.  It can be seen that Entropy+Spatial achieves higher mIoU than RAL with much less annotation budget, clearly demonstrating the effectiveness of spatial diversity in selecting informative samples.

\begin{table*}[htbp]
\centering
\caption{Comparison of alternative AL methods with a fixed budget on Cityscapes. The mean and standard deviation of 5 runs are reported.}
\begin{tabular}{lccccc}
\toprule
            & Random & BALD\cite{gal2017deep} & Entropy & RAL\cite{Casanova2020} & Entropy+Spatial \\ \midrule
Budget (\%) & 4.2    & 4.2                    & 4.2    &  14.9                    & 4.2 \\ 
mIoU (\%)   & 59.27~(0.71)& 62.85~(0.35)  & 63.09~(0.74) & 63.32~(0.93)  & \textbf{64.28}~(0.33)
 \\ \bottomrule
\end{tabular}
\label{tab:fixedbudget}
\end{table*}

\subsection{Qualitative Results}
\begin{figure*}[!htb]
	\centering
	\includegraphics[width=0.85\linewidth]{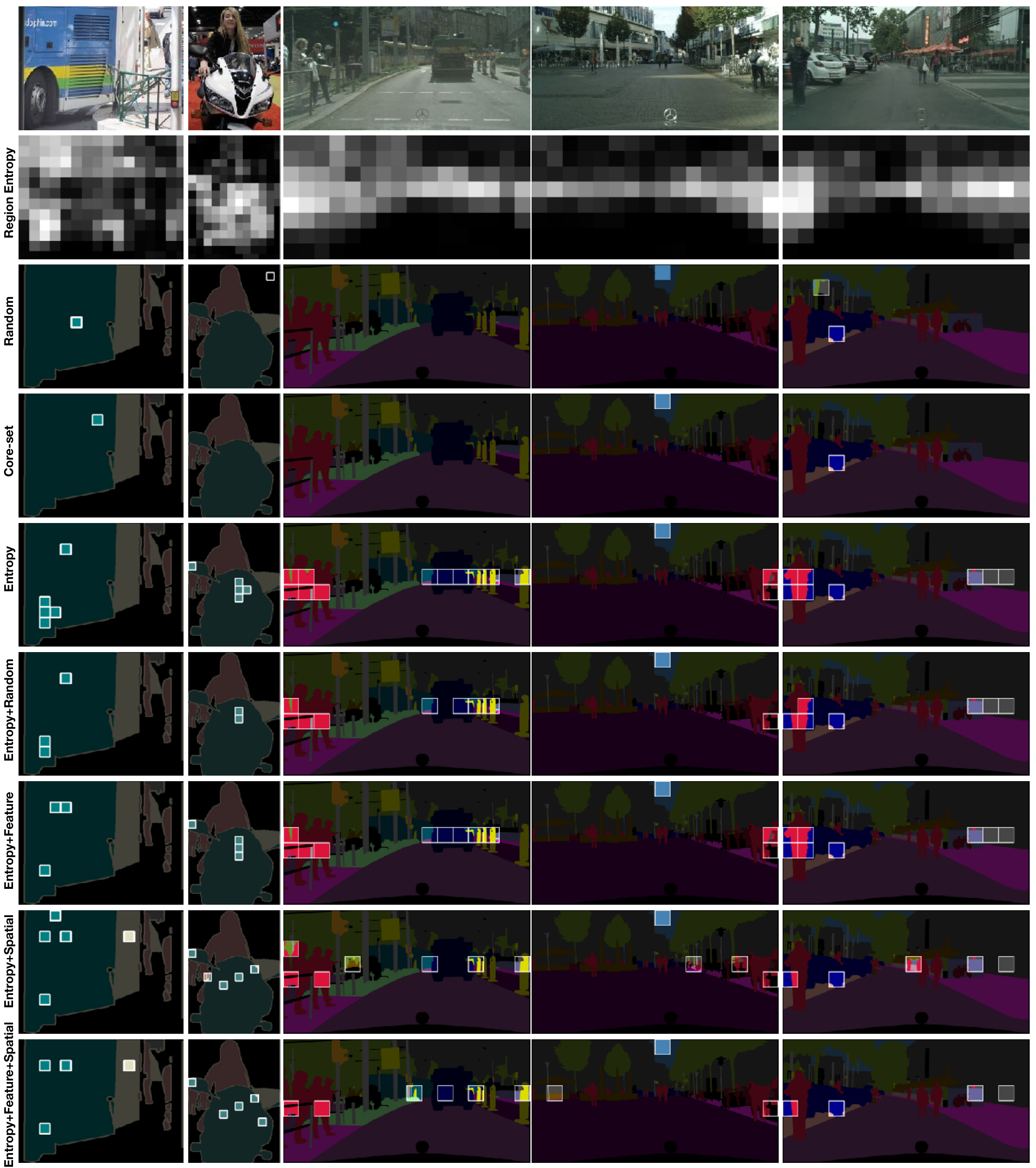}
	\caption{Visualization of regions selected by different methods at the second batch (total budget: 2k regions). Note how the incorporation of spatial diversity allows the labeling budget to be used on regions with more diverse labels and visual content.}
	\label{fig:vis_selected_region}
	\vspace{-1em}
\end{figure*}

We visualize the regions selected by different methods in Fig.~\ref{fig:vis_selected_region}. As neighboring regions usually have similar entropy values, the entropy-based method will tend to select several closely-located regions, which can be redundant for network training. Entropy+Feature is not able to effectively remedy this. By considering spatial diversity, both Entropy+Spatial and Entropy+Feature+Spatial are able to pick up regions with more diversity in terms of semantic categories (more colors in the ground-truth map). 

\begin{figure*}[!htb]
	\centering
	\includegraphics[width=0.8\linewidth]{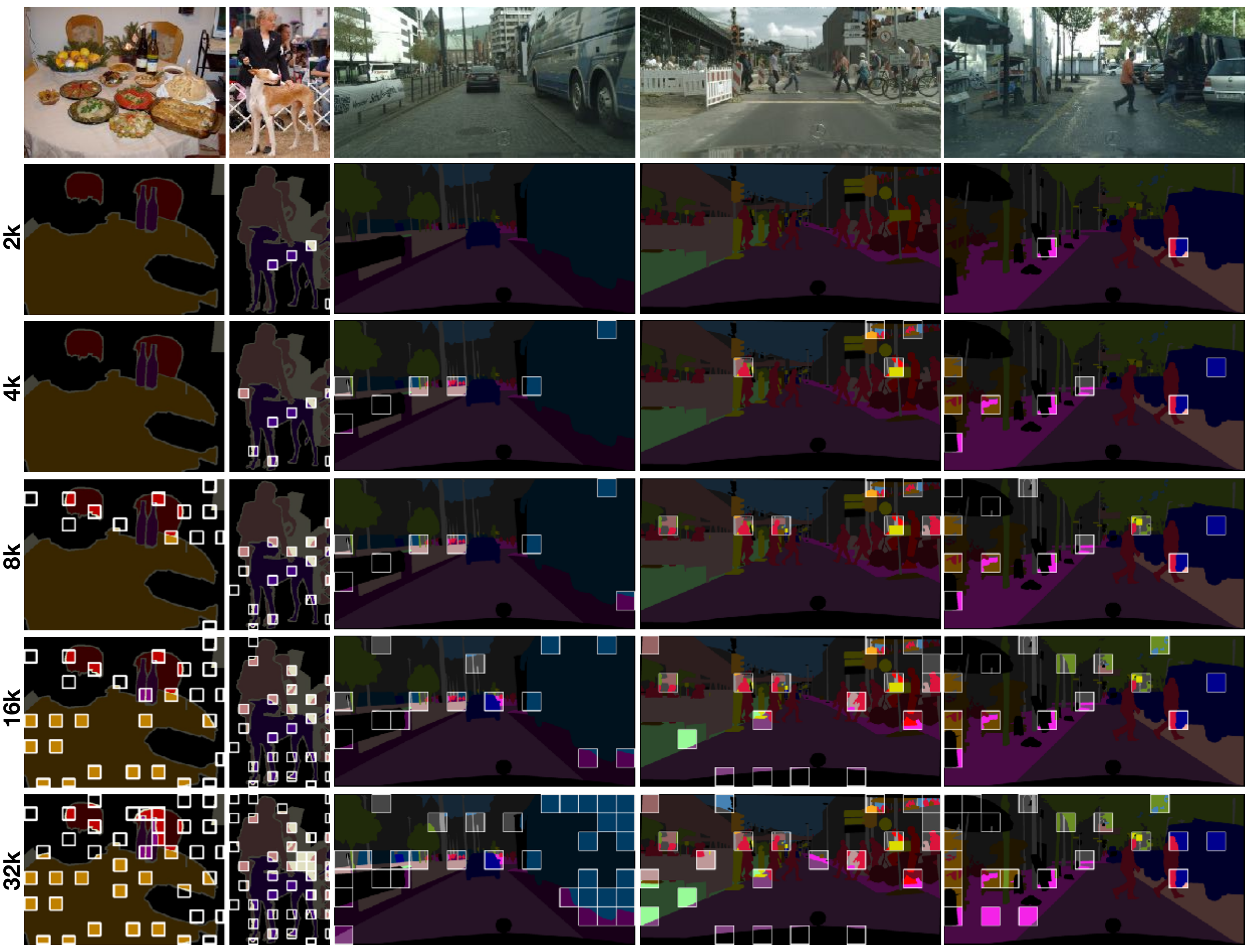}
	\caption{Visualization of selected regions of Entropy+Spatial with different annotation budgets. Note how our method allows the selection of regions that are neighbors of previously selected regions with increasing annotation budgets.}
	\label{fig:vis_selected_region_es}
    \vspace{-1em}
\end{figure*}

In Fig.~\ref{fig:vis_selected_region_es}, we present the regions selected by Entropy+Spatial at different annotation budgets to more clearly illustrate the behaviour of the iterative selection procedure. It can be seen that at the early stage of acquisition, our method will favor the selection of regions outside $\tau$ of a selected region so that the annotation budget can be used on more diverse regions. At later stages of acquisition, our algorithm will select a region if it is of high entropy even though it is a neighbor of previously selected region. Therefore, our method can flexibly handle the trade-off between uncertainty and spatial diversity and is not equivalent to a checkerboard selection pattern.

%% file: discussions.tex
In this section, we further investigate different design options for the proposed Entropy+Spatial method. This includes the form and parameters of the spatial distance function and a variant of objective function. We also provide more insight into why Entropy+Spatial outperforms  by inspecting the number of annotated images at fixed budget, computational cost for each selection strategy and the effect of feature dimension on Entropy+Feature. 

\subsection{Spatial Distance Function: Piece-wise vs. Linear}
We compare two distance measures for enforcing spatial diversity: the linear function defined in Eq.~(\ref{eq:linear_dist}) and the piece-wise function proposed in Eq.~(\ref{eq:step_dist}). We evaluate the two distance functions with the Entropy+Spatial method on both Cityscapes and PASCAL VOC~2012 and present the results in Table~\ref{tab:dist_fcn}. For Cityscapes, piece-wise function outperforms linear function for all the batches, and for PASCAL VOC~2012, linear distance function performs slightly better than piece-wise function for the first three batches but worse for the last two batches. The explanation is that linear function will enforce more spatial diversity by giving more weight to regions farther-away from a selected region, which can result in selecting a batch of spatially-diverse yet low-entropy regions. This can be beneficial for datasets consisting of mainly several dominant objects within each image (like PASCAL VOC~2012) at the early stage of training, but is harmful at later stage of training where more difficult (high-entropy) samples are desired and for datasets consisting of many small and rare categories (like Cityscapes).

\begin{table}[htb]

        \centering
        \caption{Piece-wise vs. linear function with different labeling budgets on Cityscapes and Pascal VOC~2012. Both the mean and standard deviation of 3 runs are reported.}
        
        \label{tab:dist_fcn}
        \begin{tabu} to \linewidth {X[l]X[c]X[c]X[c]X[c]X[c]}
        \toprule
        Budget  & 2k    & 4k    & 8k  & 16k & 32k\\ \midrule
        Cityscapes & \\\midrule
        Linear & 44.54~(5.56) & 53.87~(3.51) & 65.07~(0.77)   & 69.43~(0.49) & 71.59~(0.52)\\ 
        \mbox{Piece-wise}   & \textbf{51.16}~(3.49) & \textbf{62.31}~(1.15) & \textbf{67.26}~(0.35) & \textbf{70.34}~(0.50)  & \textbf{72.72}~(0.78)   \\ \midrule
        VOC~2012 & \\\midrule
        Linear & \textbf{52.48}~(0.67) & \textbf{60.05}~(0.38) & \textbf{68.59}~(0.68)   & 72.33~(0.33) & 73.85~(0.24)\\
        \mbox{Piece-wise}   & 50.98~(1.13) & 59.74~(0.89) & 68.37~(0.21) & \textbf{73.82}~(0.60)  & \textbf{74.96}~(0.32)   \\ 
        \bottomrule
        \end{tabu}
  
    \label{tab:dist_function}
    \vspace{-1em}
\end{table}

\subsection{Effect of $\tau$ on Performance}
\begin{figure}[htb]
    \centering
    \subfloat{\includegraphics[width=0.25\textwidth]{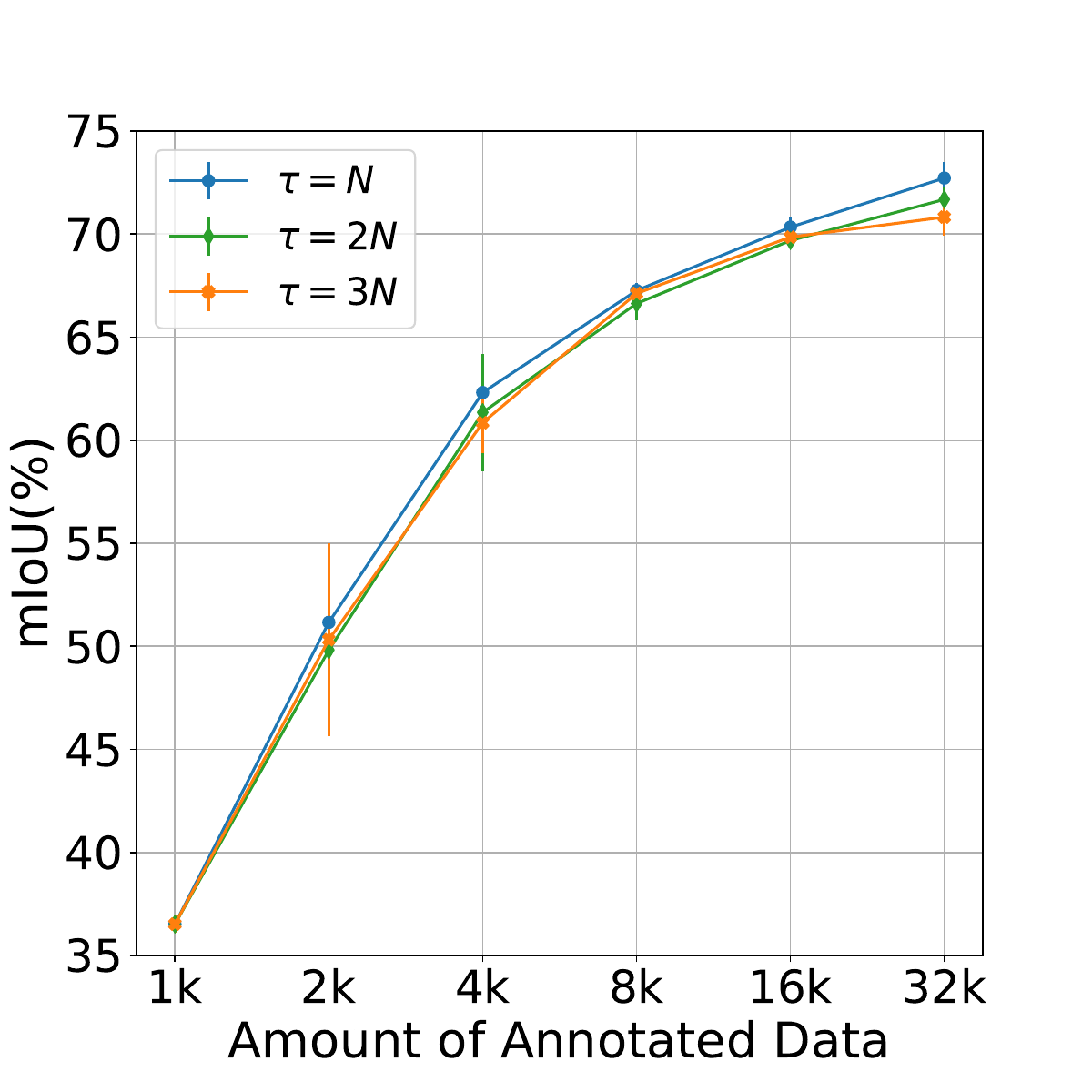}}
    \subfloat{\includegraphics[width=0.25\textwidth]{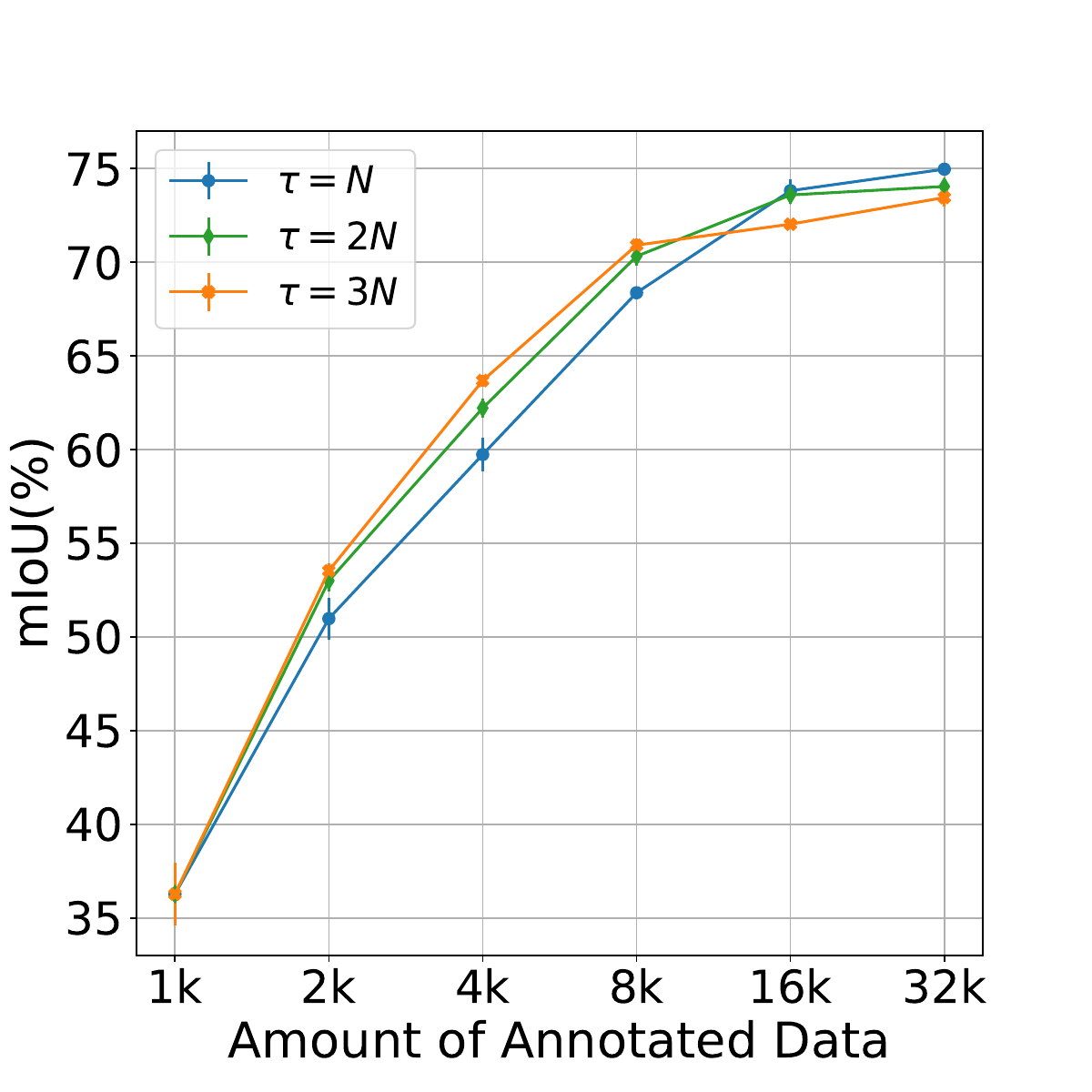}}
 
    \caption{Effect of $\tau$ on performance for Cityscapes (left) and PASCAL VOC~2012 (right).}
    \label{fig:senstivity_tau}
    \vspace{-1em}
\end{figure}

The parameter of $\tau$ in Eq.~(\ref{eq:step_dist}) essentially defines a cut-off for which region is considered close to a selected region and the selection of which should be discouraged. A larger $\tau$ will enforce more spatial diversity. We investigate the effect of various $\tau$ values and the results are presented in Fig.~\ref{fig:senstivity_tau}. It can be seen that on Cityscapes (region size $N=128$), larger $\tau$ degrades performance across the curve, and on PASCAL VOC~2012 (region size $N=32$), using a large $\tau$ is beneficial for the first 4 batches, but degrades performance for the last 2 batches. We analyze the reasons for this behaviour as follows. Firstly, the PASCAL VOC~2012 dataset mostly consists of images with only a few dominant objects. Using larger $\tau$ increases the diversity of selected samples and helps to train a better model when the labeling budget is limited. When the labeling budget increases, a large $\tau$ may negatively affect the model as it may disregard those regions that the model is less confident in simply because they are close by. In contrast, the Cityscapes dataset consists of images each with many semantic objects, and some objects like traffic lights and signs, are often under-represented. By overly enforcing the spatial diversity objective, the selected regions may miss those small and less frequent categories. Thus, we observe worse performance with increasing $\tau$ on Cityscapes.  

\subsection{Effect of $a$, $b$ and $c$ on Performance}
\begin{figure}[htb]
    \centering
    \subfloat{\includegraphics[width=0.25\textwidth]{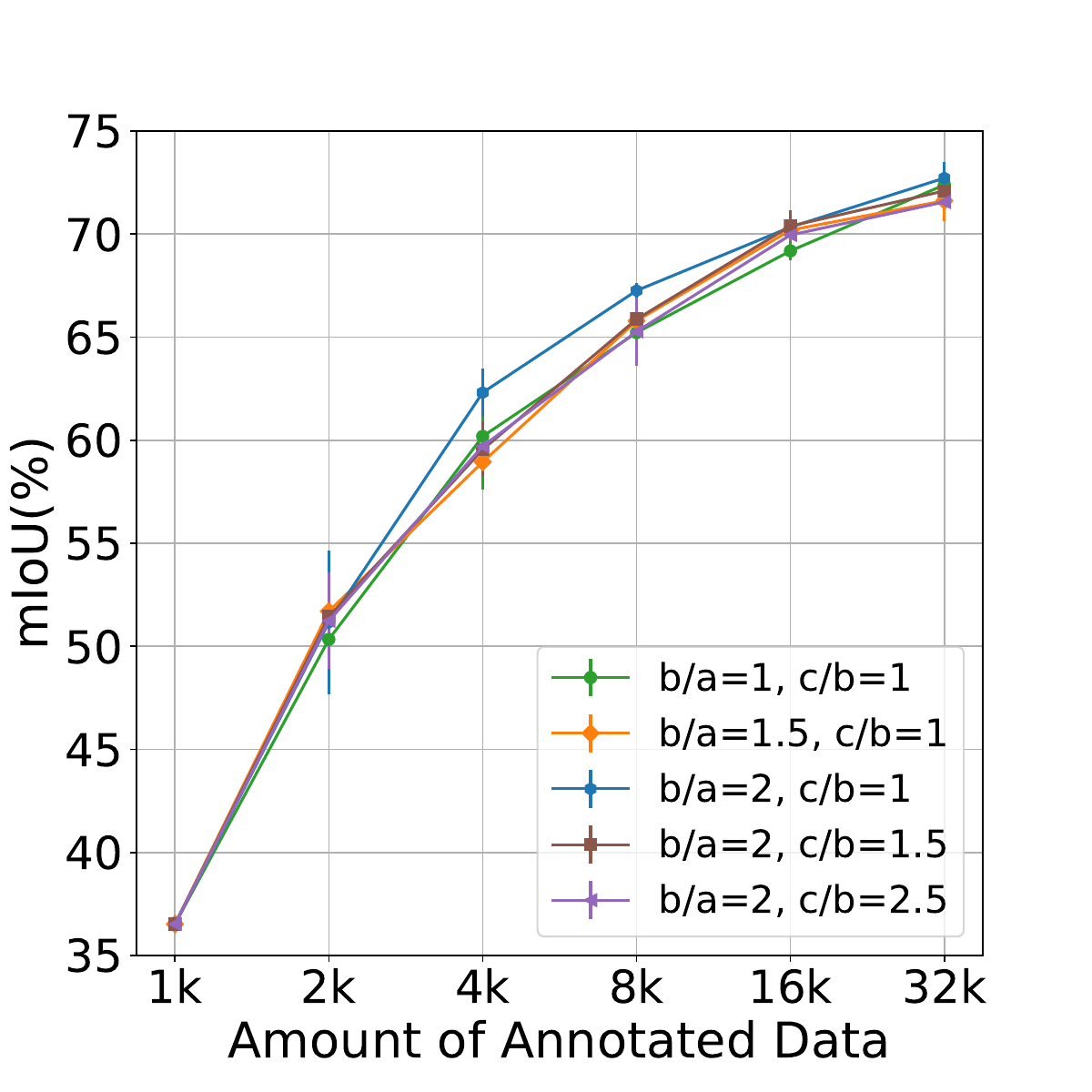}}
    \subfloat{\includegraphics[width=0.25\textwidth]{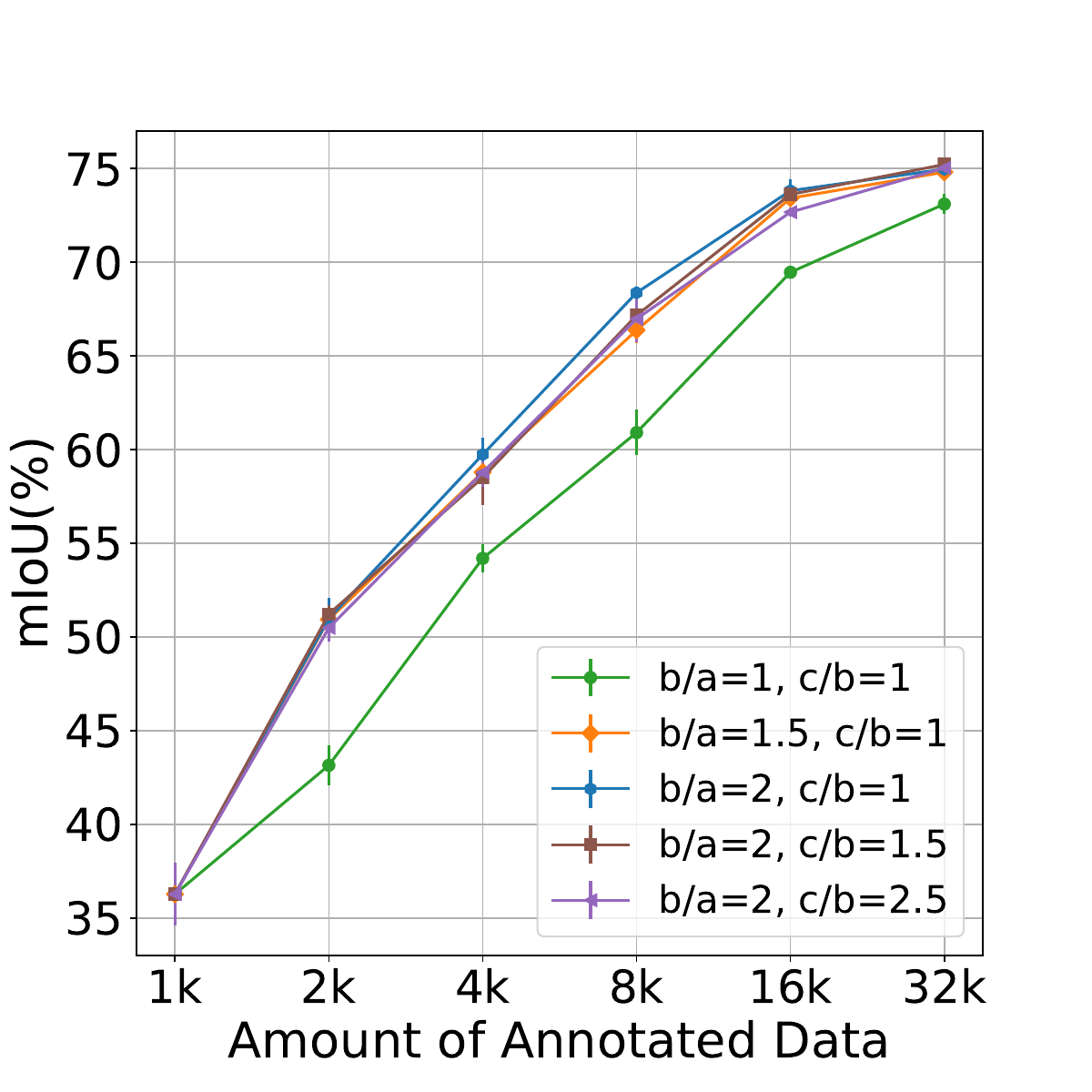}}
    
	\caption{Sensitivity of $a,b,c$ on performance for Cityscapes (left) and PASCAL VOC~2012 (right).}
    \label{fig:senstivity_a_b_c}
    \vspace{-1em}
\end{figure}

For the piece-wise distance function defined in Eq.~(\ref{eq:step_dist}), parameters $a$, $b$ and $c$ specify the value for each piece. As each term in Eq.~(\ref{eq:obj_maxmin}) is normalized before summation, only their ratios matter (i.e., $b/a$ and $c/b$). Also, for the spatial distance to be a metric, the ratios have to satisfy $1 \leq b/a \leq 2$ and $c/b \geq 1$ (proof is in Appendix A.). The ratio $b/a$ determines the relative weight assigned to regions within vs. beyond $\tau$ to a specific region, while $c/b$ determine the relative weight assigned to regions from the same vs. from different images. To investigate the effect of $b/a$, we keep $c/b=1$ and vary $b/a$. Similarly, to investigate the effect of $c/b$, we keep $b/a=2$ and vary $c/b$. The experiment is run with Entropy+Spatial and the results are presented in Fig.~\ref{fig:senstivity_a_b_c}. It can be seen that when $c/b$ is fixed, increasing $b/a$ from 1 to 2 leads to better performance, indicating that weighting more on regions outside the cut-off $\tau$ helps to select better samples for training. When $b/a$ is fixed, increasing $c/b$ from 1 to 2.5 does not have significant effect on performance, indicating that it is unnecessary to weight more on regions from a different image.

\subsection{Effect of $N$ on Performance}
\begin{figure}[htb]
  \centering
  \subfloat{\includegraphics[width=0.25\textwidth]{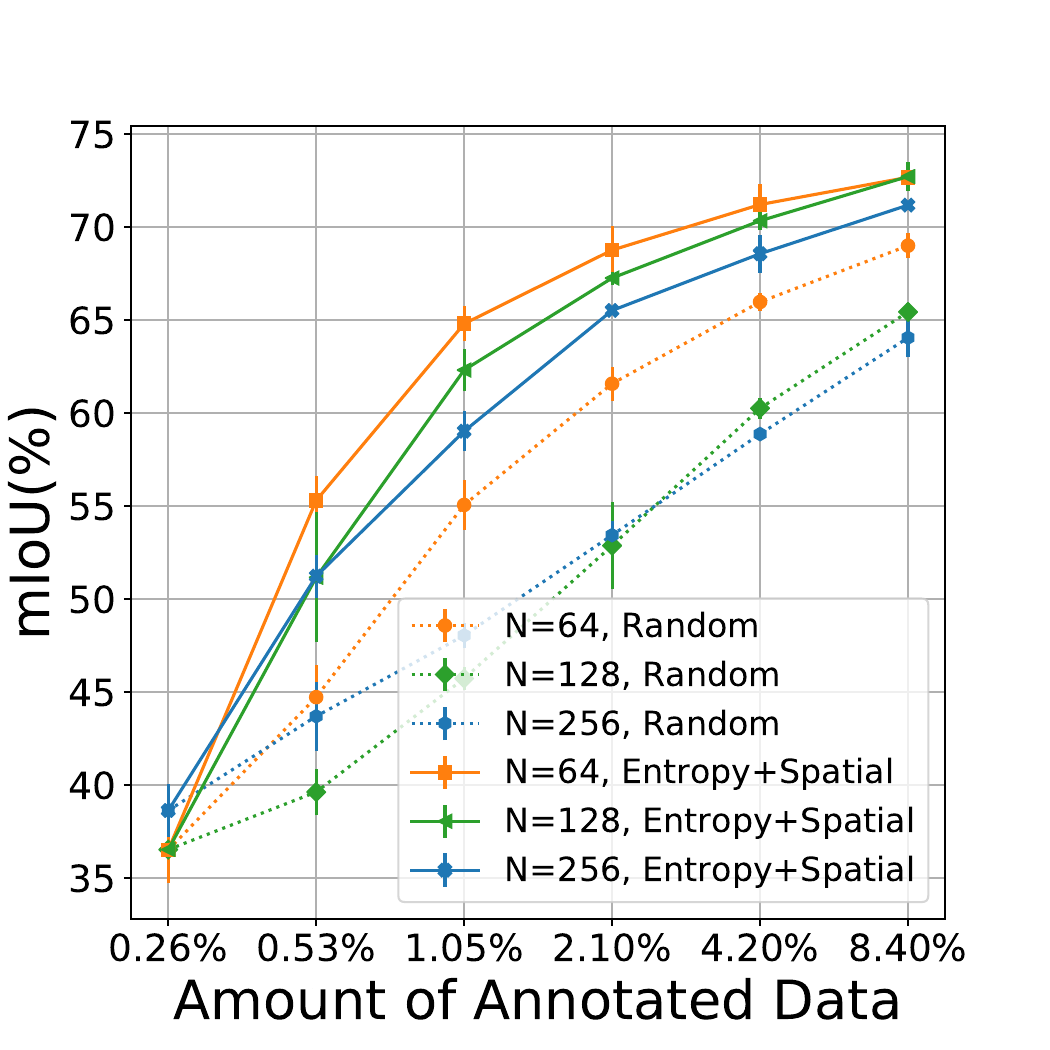}}
  \subfloat{\includegraphics[width=0.25\textwidth]{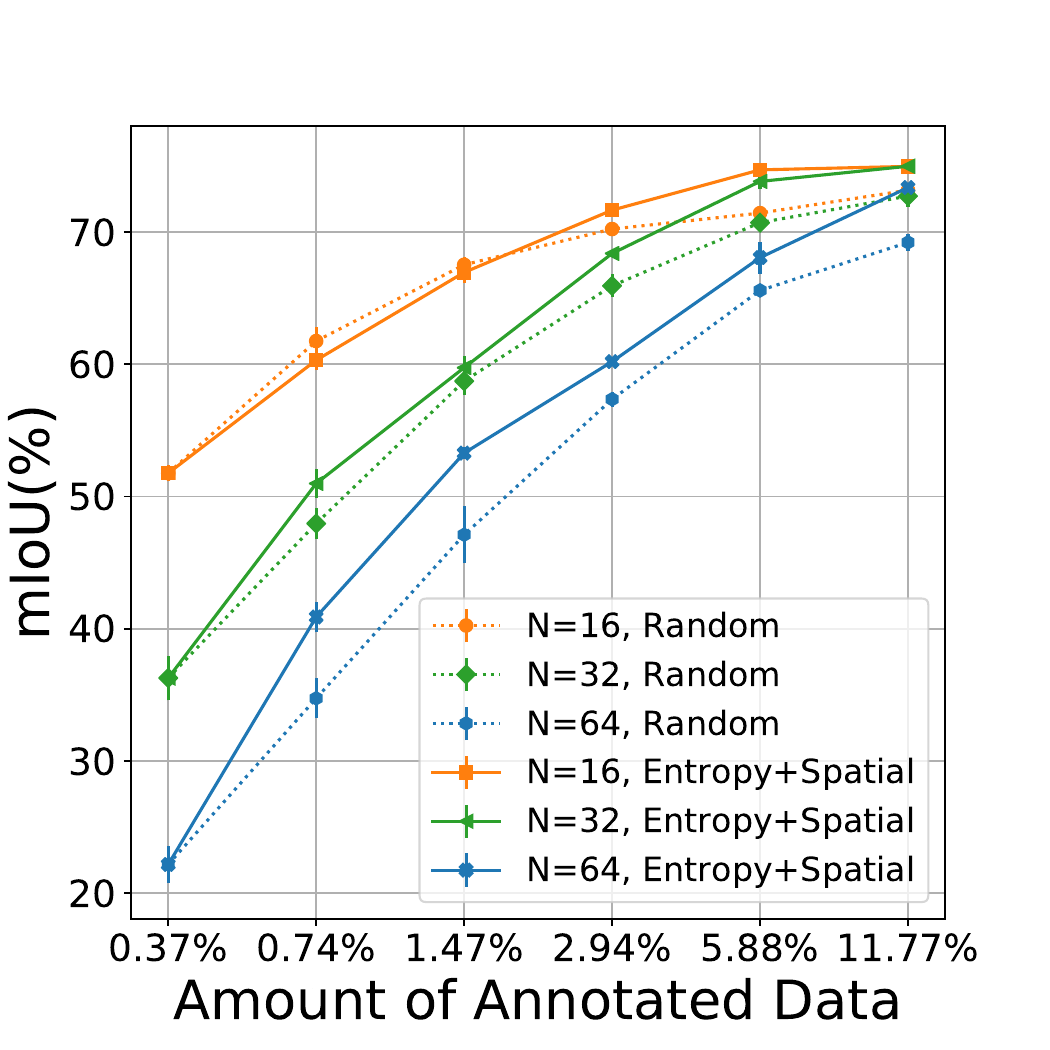}}
   
  \caption{Effect of $N$ on performance for Cityscapes (left) and PASCAL VOC~2012 (right).}
  \label{fig:effect_N}
  \vspace{-1em}
\end{figure}

To study the effect of $N$ (region size), we keep all other parameters the same, i.e., $\tau=N, a=1, b=c=2$. Also, the amount of pixels selected at each batch is kept fixed, and only the value of $N$ varies. For example, at batch 1, we select 1000 regions for $N=128$, 250 regions for $N=256$, and 4000 regions for $N=64$. The results are shown in Fig.~\ref{fig:effect_N}. It can be seen that for both the Random baseline and the proposed Entropy+Spatial method, smaller region size leads to better performance as it allows the annotation budget to be allocated to label more diverse content. 

\subsection{Diversity Set Function: Max-Min vs. Max-Sum}
We further evaluate an alternative set function for region-based AL. As described in Section~\ref{sect:uni_obj}, there is an alternative choice of set functions $\Omega$ and $\Phi$ to both be summation, which is adopted by USDM \cite{YangMNCH15}, and that we denote as \textit{Max-Sum}. As the optimization is NP-hard, we adopted the 2-approximation algorithm proposed in \cite{borodin2012max} to solve it greedily. As shown in Table~\ref{tab:maxmin_vs_maxsum}, the \textit{Max-Min} formulation performs consistently better than \textit{Max-Sum} for all batches on both Cityscapes and PASCAL VOC~2012. The superiority of the \textit{Max-Min} formulation is likely due to the more strict diversity constraint, as \textit{Min} is more strict than \textit{Sum} in assessing a candidate region's closeness to all previously selected samples. 

\begin{table}[htb]
        \centering
        \caption{Comparing \textit{Max-Min} and \textit{Max-Sum} as objective with Entropy+Spatial on Cityscapes and PASCAL VOC~2012. Both the mean and standard deviation of 3 runs are reported.}
        
        \label{tab:maxmin_vs_maxsum}
        \begin{tabu} to \linewidth {X[l]X[c]X[c]X[c]X[c]X[c]}
        \toprule
        Budget  & 2k    & 4k    & 8k  & 16k & 32k\\ \midrule
        Cityscapes & \\\midrule
        \mbox{\textit{Max-Sum}} & 48.45~(3.73) & 59.49~(0.41) & 66.14~(1.11)   & 68.65~(0.38) & 71.98~(0.58)\\ 
        \mbox{\textit{Max-Min}}  & \textbf{51.16}~(3.49) & \textbf{62.31}~(1.15) & \textbf{67.26}~(0.35) & \textbf{70.34}~(0.50)  & \textbf{72.72}~(0.78)   \\ \midrule
        VOC~2012 & \\\midrule
        \mbox{\textit{Max-Sum}} & 44.80~(0.38) & 52.33~(0.45) & 59.81~(1.32)   & 68.30~(0.63) & 72.75~(0.33)\\
        \mbox{\textit{Max-Min}}   & \textbf{50.98}~(1.13) & \textbf{59.74}~(0.89) & \textbf{68.37}~(0.21) & \textbf{73.82}~(0.60)  & \textbf{74.96}~(0.32)   \\ 
        \bottomrule
        \end{tabu}
        \vspace{-1em}
\end{table}

\subsection{Number of Annotated Images at Fixed Budget}
To give some intuition on the behavior of the various selection methods, we show the number of images for which some regions have been annotated given a fixed annotation budget (i.e., fixed number of total selected regions) in Table~\ref{tab:num_anno_image}. It can be seen that given a fixed budget, the Random baseline spreads annotated regions over the most images while Entropy covers the least, implying that Random can select a diverse set of samples, but the selected samples may not be informative. On the other hand, Entropy can select high-uncertainty samples, but the selected samples may be redundant as it tends to select neighboring regions from the same image (refer to  Fig.~\ref{fig:vis_selected_region} for illustrated examples). The number of images selected by Entropy+Spatial (ES) is larger than Entropy and Entropy+Feature (EF), but smaller than Random, implying that it can select more diverse samples (prefer to select regions from different images) than Entropy and Entropy+Feature, and use less images than Random.
\begin{table}[htb]
        \centering
        \caption{Number of annotated images at fixed budget. Both the mean and standard deviation of 3 runs are reported.}
        
        \label{tab:num_anno_image}
        \begin{tabu} to \linewidth {X[l]X[c]X[c]X[c]X[c]X[c]}
        \toprule
        Budget  & 2k    & 4k    & 8k  & 16k & 32k\\ \midrule
        \mbox{Random} & 1460~(6) & 2197~(4) & 2784~(5) & 2964~(1) & 2975~(0) \\
        \mbox{Entropy}  & 1211~(15) & 1689~(18) & 2275~(12)   & 2723~(6) & 2930~(3)\\ 
        \mbox{EF} & 1243~(22) & 1721~(33) & 2288~(21) & 2723~(14) & 2941~(9) \\
        \mbox{ES} & 1362~(26) & 1983~(20) & 2612~(16) & 2931~(4) & 2975~(1) \\
  
        \bottomrule
        \end{tabu}
        \vspace{-1em}
\end{table}

\subsection{Computational Cost Analysis}
Here we analyze the computational cost required to evaluate the active selection objectives during each greedy selection iteration of Algorithm 1. The majority time is spent on distance computation. In implementation, we maintain a vector to store the minimum distance between each sample in the dataset and currently selected pool $\set{L}_t\cup\set{B}_t$, and update this vector with every newly selected sample. As such, the total times of distance computation is $O(n\cdot K)$, where $n$ is the total number of samples in the dataset, and $K$ is the total number of selected samples for a batch.  For Entropy+Feature, the time complexity is $O(n\cdot K\cdot f_{dim})$,  where $f_{dim}$ is the dimension of the feature vector. The time complexity for Entropy+Spatial is $O(n\cdot K)$, which is more efficient as it does not involve high-dimensional vector computation. Furthermore, Entropy+Spatial can reuse the distance vector computed in the previous iteration while Entropy+Feature needs to recompute the distance at each iteration as the features are updated (refer to Section~\ref{sect:imp} for more details). Table~\ref{tab:comp_cost} shows the running time for the different selection methods on Cityscapes. It can be seen that Entropy+Spatial is 20x faster than Entropy+Feature.
\begin{table}[htb]
        \centering
        \caption{Running time measured in seconds for different selection methods at various budgets on Cityscapes. We use a CUDA implementation to compute feature and spatial distance. The experimental platform is a DGX-1 with Tesla V100 cards. The mean and standard deviation of 3 runs are reported.}
        
        \label{tab:comp_cost}
        \begin{tabu} to \linewidth {X[l]X[c]X[c]X[c]X[c]X[c]}
        \toprule
        Budget  & 2k    & 4k    & 8k  & 16k & 32k\\ \midrule
        \mbox{Entropy}  & 3.92~(0.53) & 7.90~(1.46) & 13.5~(0.22)   & 29.6~(2.74) & 71.6~(9.56)\\ 
        \mbox{EF} & 398~(56.5) & 824~(90.3) & 1784~(62.9) & 3273~(168) & 6350~(629) \\
        \mbox{ES} & 32.0~(5.90) & 41.1~(8.40) & 81.9~(6.08) & 157~(6.50) & 322~(12.0) \\
  
        \bottomrule
        \end{tabu}
        \vspace{-1em}
\end{table}

\subsection{Effect of Feature Dimension on Entropy+Feature}
\begin{figure}[htb]
    \centering
    \subfloat{\includegraphics[width=0.25\textwidth]{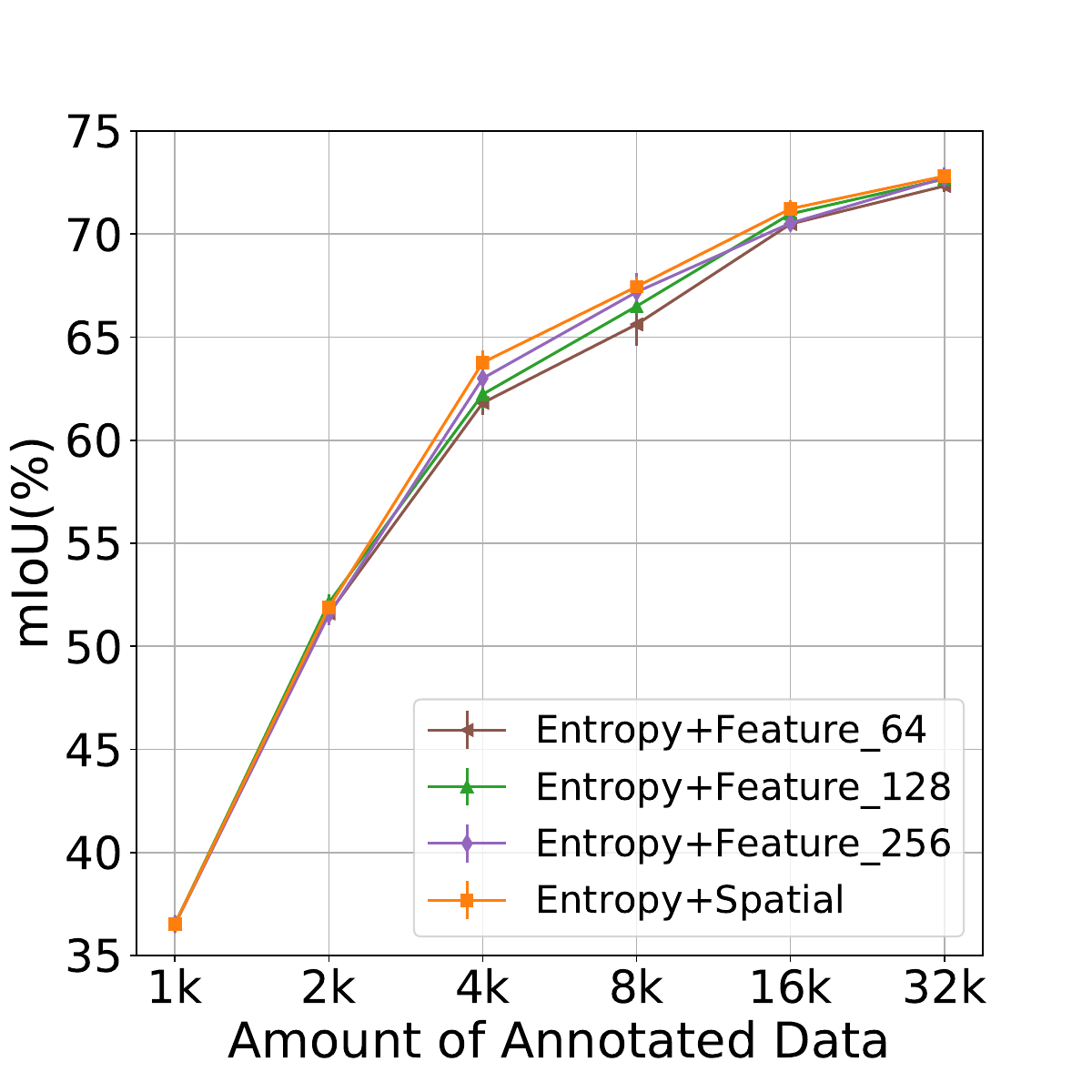}}
    \subfloat{\includegraphics[width=0.25\textwidth]{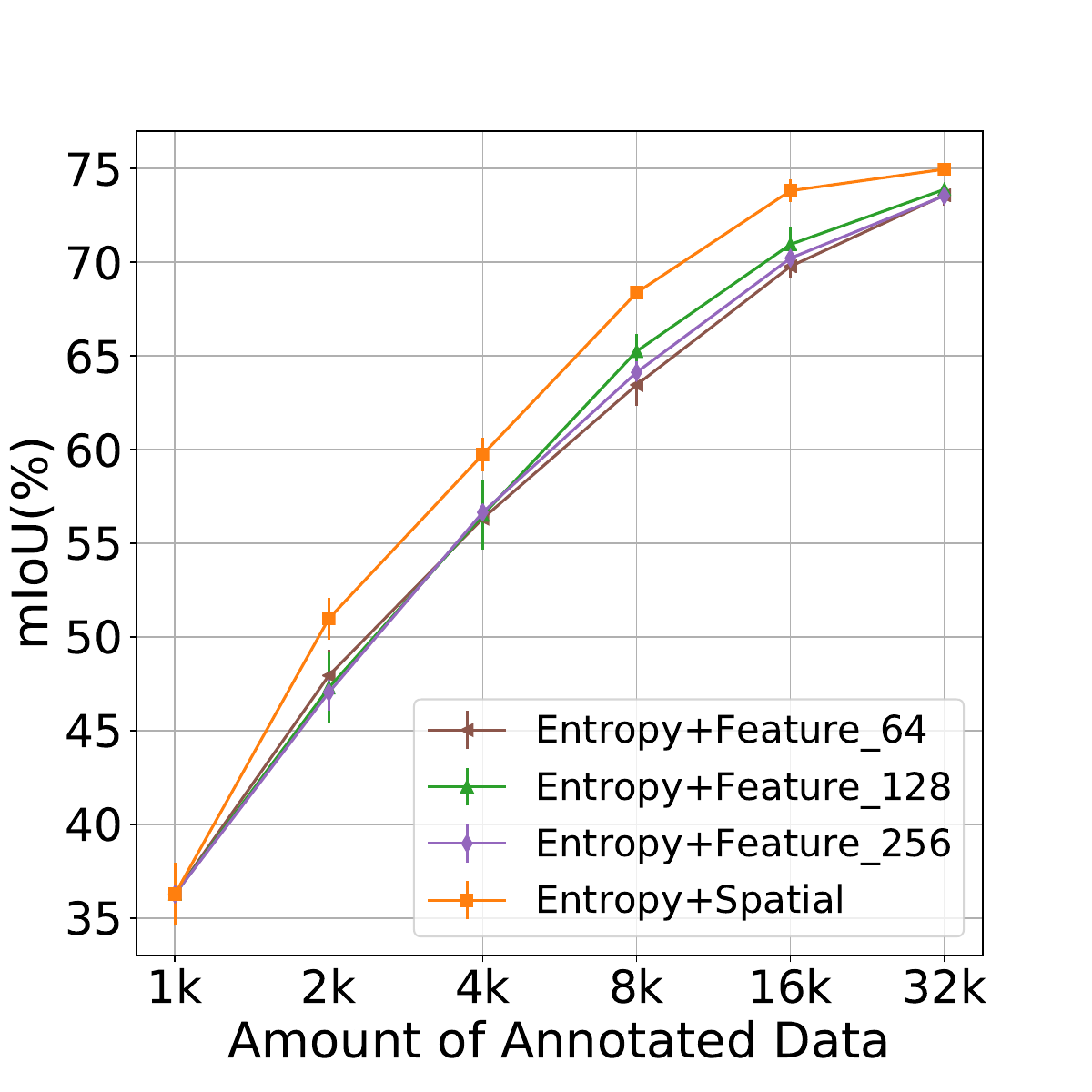}}
    
	\caption{Effect of feature dimension on Entropy+Feature for Cityscapes (left) and PASCAL VOC~2012 (right).}
    \label{fig:feat_dim_EF}
    \vspace{-1em}
\end{figure}

We compare Entropy+Spatial with Entropy+Feature using different feature dimensions to further demonstrate the advantage of spatial diversity over feature diversity. The original feature vector has a dimension of 256 (refer to Section~\ref{sect:imp} for more details). We use PCA to project the feature vector to a dimension of 64,128 and 256 respectively and the results are presented in Fig.~\ref{fig:feat_dim_EF}. It can be seen that increasing or decreasing the feature dimension does not affect the performance of Entropy+Feature significantly, and Entropy+Spatial still outperforms consistently.

%% file: conclusions.tex
In this work, we formulated the region-based active learning problem as a Max-Min diversification problem and proposed a greedy algorithm to solve it efficiently. We introduced spatial diversity as a new objective for optimization and defined spatial distance of two regions as a piece-wise function to encourage \emph{local spatial diversity}. Extensive experiments on Cityscapes and PASCAL VOC~2012 demonstrated that the proposed spatial diversity objective can facilitate selecting label-diverse regions which are known to be critical for the training of deep neural network models. Our diversity objective complements existing uncertainty and feature diversity measures for region-based AL. Compared with the traditional feature diversity objective, spatial diversity has the benefit of not relying on a good feature extractor and does not suffer from the p-norm limitation. The proposed Entropy+Spatial method obtains state-of-the-art results for active semantic segmentation, achieving $95\%$ performance of fully supervised methods with only $8.4\%$ and $5.9\%$ pixels labeled on Cityscapes and PASCAL VOC~2012, respectively.

%% file: supplemental.tex
\section{Distance Metric}

\begin{theorem}
$d_s(x,y)$ defined in Eq.~(5) is a metric if $c\geq b\geq a > 0$ and $b\leq 2a$.
\end{theorem}

Proof:
First, we note that for $a,b,c$ satisfying $c\geq b\geq a > 0$ we have for all $x,y$ that $d_s(x,y) \geq 0$ and $d_s(x,y)=d_s(y,x)$ directly from the definition of $d_s$; by definition we also have $d_s(x,y) = 0$ if and only if $x = y$. Thus
we just need to prove the triangle inequality $d_s(x, y) \leq d_s(x, z) + d_s(z, y)$ for any 3 regions $x,y,z$. Since $d_s(x,y)$ can only take 4 values: $0, a, b, c$, we verify that the inequality holds in each of these cases:
\begin{enumerate}
 \item If $d_s(x, y) = 0$, because $d_s(x,z)$ and $d_s(z,y)$ are non-negative, the triangle inequality holds.
 \item If $d_s(x, y) = a$, a violation can only happen if both $d_s(x, z)$ and $d_s(z,y)$ are 0. But this is a contradiction as this implies $x=y=z$ and $d_s(x,y)=0$. Thus the triangle inequality holds in this case.
 \item If $d_s(x, y) = b$, a violation could occur if $d_s(x,z) = d_s(z,y) = 0$, $d_s(x,z) = 0$ and $ d_s(z,y)=a$, $d_s(x,z) = a$ and $d_s(z,y)=0$ or $d_s(x,z)=a$ and $d_s(z,y)=a$. We analyze each of these 4 possibilities. First, if both $d_s(x,z) = d_s(z,y) = 0$, this is a contradiction as in the previous case 2. If $d_s(x,z) = 0$, it implies $x=z$ and $d_s(x,y)=d_s(z,y)=a$ which contradicts the premise $d_s(x,y)=b$; a similar argument applies to the third possibility. Finally, for the fourth possibility, given $b\leq 2a$, we have $d_s(x,y)=b\leq d_s(x,z)+d_s(z,y)=2a$ and the triangle inequality holds.
 \item If $d_s(x, y) = c$, then $x$ and $y$ are two regions in different images, so at least one of $d_s(x,z)$ or $d_s(z,y) = c$, since otherwise $z$ has to be in both images containing $x$ and $y$ which is impossible. Thus the triangle inequality holds.
\end{enumerate}

%% file: main.bbl
\begin{thebibliography}{10}
\providecommand{\url}[1]{#1}
\csname url@samestyle\endcsname
\providecommand{\newblock}{\relax}
\providecommand{\bibinfo}[2]{#2}
\providecommand{\BIBentrySTDinterwordspacing}{\spaceskip=0pt\relax}
\providecommand{\BIBentryALTinterwordstretchfactor}{4}
\providecommand{\BIBentryALTinterwordspacing}{\spaceskip=\fontdimen2\font plus
\BIBentryALTinterwordstretchfactor\fontdimen3\font minus
  \fontdimen4\font\relax}
\providecommand{\BIBforeignlanguage}[2]{{%
\expandafter\ifx\csname l@#1\endcsname\relax
\typeout{** WARNING: IEEEtran.bst: No hyphenation pattern has been}%
\typeout{** loaded for the language `#1'. Using the pattern for}%
\typeout{** the default language instead.}%
\else
\language=\csname l@#1\endcsname
\fi
#2}}
\providecommand{\BIBdecl}{\relax}
\BIBdecl

\bibitem{settles.book12}
B.~Settles, \emph{Active Learning}, ser. Synthesis Lectures on Artificial
  Intelligence and Machine Learning.\hskip 1em plus 0.5em minus 0.4em\relax
  Morgan \& Claypool, 2012.

\bibitem{YangMNCH15}
Y.~Yang, Z.~Ma, F.~Nie, X.~Chang, and A.~G. Hauptmann, ``Multi-class active
  learning by uncertainty sampling with diversity maximization,''
  \emph{International Journal of Computer Vision}, 2015.

\bibitem{gal2017deep}
Y.~Gal, R.~Islam, and Z.~Ghahramani, ``Deep bayesian active learning with image
  data,'' in \emph{ICML}, 2017.

\bibitem{SenerS18}
O.~Sener and S.~Savarese, ``Active learning for convolutional neural networks:
  {A} core-set approach,'' in \emph{ICLR}, 2018.

\bibitem{Cordts2016Cityscapes}
M.~Cordts, M.~Omran, S.~Ramos, T.~Rehfeld, M.~Enzweiler, R.~Benenson,
  U.~Franke, S.~Roth, and B.~Schiele, ``The cityscapes dataset for semantic
  urban scene understanding,'' in \emph{CVPR}, 2016.

\bibitem{MackowiakLGDLR18}
R.~Mackowiak, P.~Lenz, O.~Ghori, F.~Diego, O.~Lange, and C.~Rother, ``{CEREALS}
  - cost-effective region-based active learning for semantic segmentation,'' in
  \emph{BMVC}, 2018.

\bibitem{Casanova2020}
A.~Casanova, P.~Pinheiro, N.~Rostamzadeh, and P.~Christopher, ``{Reinforced
  Active Learning for Image Segmentation},'' in \emph{ICLR}, 2020.

\bibitem{LewisG94}
D.~D. Lewis and W.~A. Gale, ``A sequential algorithm for training text
  classifiers,'' in \emph{Proceedings of the 17th Annual International
  {ACM-SIGIR} Conference on Research and Development in Information Retrieval.
  Dublin, Ireland, 3-6 July 1994 (Special Issue of the {SIGIR} Forum)}, 1994.

\bibitem{JoshiPP09}
A.~J. Joshi, F.~Porikli, and N.~Papanikolopoulos, ``Multi-class active learning
  for image classification,'' in \emph{CVPR}, 2009.

\bibitem{houlsby2011bayesian}
N.~Houlsby, F.~Husz{\'a}r, Z.~Ghahramani, and M.~Lengyel, ``Bayesian active
  learning for classification and preference learning,'' \emph{arXiv preprint
  arXiv:1112.5745}, 2011.

\bibitem{NguyenS04}
H.~T. Nguyen and A.~W.~M. Smeulders, ``Active learning using pre-clustering,''
  in \emph{ICML}, 2004.

\bibitem{WangY13}
Z.~Wang and J.~Ye, ``Querying discriminative and representative samples for
  batch mode active learning,'' in \emph{ACM SIGKDD}, 2013.

\bibitem{VezhnevetsBF12}
A.~Vezhnevets, J.~M. Buhmann, and V.~Ferrari, ``Active learning for semantic
  segmentation with expected change,'' in \emph{CVPR}, 2012.

\bibitem{BeluchGNK18}
W.~H. Beluch, T.~Genewein, A.~N{\"{u}}rnberger, and J.~M. K{\"{o}}hler, ``The
  power of ensembles for active learning in image classification,'' in
  \emph{CVPR}, 2018.

\bibitem{Yoo2019}
D.~Yoo and I.~S. Kweon, ``Learning loss for active learning.'' in \emph{CVPR},
  2019, pp. 93--102.

\bibitem{ZhouSZGGL17}
Z.~Zhou, J.~Y. Shin, L.~Zhang, S.~R. Gurudu, M.~B. Gotway, and J.~Liang,
  ``Fine-tuning convolutional neural networks for biomedical image analysis:
  Actively and incrementally,'' in \emph{CVPR}, 2017.

\bibitem{dutt2016active}
S.~Dutt~Jain and K.~Grauman, ``Active image segmentation propagation,'' in
  \emph{CVPR}, 2016.

\bibitem{YangZCZC17}
L.~Yang, Y.~Zhang, J.~Chen, S.~Zhang, and D.~Z. Chen, ``Suggestive annotation:
  {A} deep active learning framework for biomedical image segmentation,'' in
  \emph{Medical Image Computing and Computer Assisted Intervention - {MICCAI}
  2017 - 20th International Conference, Quebec City, QC, Canada, September
  11-13, 2017, Proceedings, Part {III}}, 2017, pp. 399--407.

\bibitem{mahapatra2018efficient}
D.~Mahapatra, B.~Bozorgtabar, J.-P. Thiran, and M.~Reyes, ``Efficient active
  learning for image classification and segmentation using a sample selection
  and conditional generative adversarial network,'' in \emph{International
  Conference on Medical Image Computing and Computer-Assisted Intervention},
  2018.

\bibitem{sinha2019variational}
S.~Sinha, S.~Ebrahimi, and T.~Darrell, ``Variational adversarial active
  learning,'' in \emph{Proceedings of the IEEE International Conference on
  Computer Vision}, 2019, pp. 5972--5981.

\bibitem{kasarla2019region}
T.~Kasarla, G.~Nagendar, G.~M. Hegde, V.~Balasubramanian, and C.~Jawahar,
  ``Region-based active learning for efficient labeling in semantic
  segmentation,'' in \emph{2019 IEEE Winter Conference on Applications of
  Computer Vision (WACV)}.\hskip 1em plus 0.5em minus 0.4em\relax IEEE, 2019,
  pp. 1109--1117.

\bibitem{siddiqui2020viewal}
Y.~Siddiqui, J.~Valentin, and M.~Nie{\ss}ner, ``Viewal: Active learning with
  viewpoint entropy for semantic segmentation,'' in \emph{Proceedings of the
  IEEE/CVF Conference on Computer Vision and Pattern Recognition}, 2020, pp.
  9433--9443.

\bibitem{ravi1994heuristic}
S.~S. Ravi, D.~J. Rosenkrantz, and G.~K. Tayi, ``Heuristic and special case
  algorithms for dispersion problems,'' \emph{Operations Research}, 1994.

\bibitem{gollapudi2009axiomatic}
S.~Gollapudi and A.~Sharma, ``An axiomatic approach for result
  diversification,'' in \emph{Proceedings of the 18th international conference
  on World wide web}, 2009.

\bibitem{ren2020survey}
P.~Ren, Y.~Xiao, X.~Chang, P.-Y. Huang, Z.~Li, X.~Chen, and X.~Wang, ``A survey
  of deep active learning,'' \emph{arXiv preprint arXiv:2009.00236}, 2020.

\bibitem{pascal-voc-2012}
M.~Everingham, L.~Van~Gool, C.~K.~I. Williams, J.~Winn, and A.~Zisserman, ``The
  {PASCAL} {V}isual {O}bject {C}lasses {C}hallenge 2012 {(VOC2012)}
  {R}esults,''
  http://www.pascal-network.org/challenges/VOC/voc2012/workshop/index.html.

\bibitem{liu2019auto}
C.~Liu, L.-C. Chen, F.~Schroff, H.~Adam, W.~Hua, A.~L. Yuille, and L.~Fei-Fei,
  ``Auto-deeplab: Hierarchical neural architecture search for semantic image
  segmentation,'' in \emph{CVPR}, 2019.

\bibitem{deeplabv3plus2018}
L.-C. Chen, Y.~Zhu, G.~Papandreou, F.~Schroff, and H.~Adam, ``Encoder-decoder
  with atrous separable convolution for semantic image segmentation,'' in
  \emph{ECCV}, 2018.

\bibitem{Chollet17}
F.~Chollet, ``Xception: Deep learning with depthwise separable convolutions,''
  in \emph{2017 {IEEE} Conference on Computer Vision and Pattern Recognition,
  {CVPR} 2017, Honolulu, HI, USA, July 21-26, 2017}, 2017, pp. 1800--1807.

\bibitem{deng2009imagenet}
J.~Deng, W.~Dong, R.~Socher, L.-J. Li, K.~Li, and L.~Fei-Fei, ``Imagenet: A
  large-scale hierarchical image database,'' in \emph{2009 IEEE conference on
  computer vision and pattern recognition}.\hskip 1em plus 0.5em minus
  0.4em\relax Ieee, 2009, pp. 248--255.

\bibitem{lin2017feature}
T.-Y. Lin, P.~Doll{\'a}r, R.~Girshick, K.~He, B.~Hariharan, and S.~Belongie,
  ``Feature pyramid networks for object detection,'' in \emph{Proceedings of
  the IEEE conference on computer vision and pattern recognition}, 2017, pp.
  2117--2125.

\bibitem{richter2016playing}
S.~R. Richter, V.~Vineet, S.~Roth, and V.~Koltun, ``Playing for data: Ground
  truth from computer games,'' in \emph{European conference on computer
  vision}.\hskip 1em plus 0.5em minus 0.4em\relax Springer, 2016, pp. 102--118.

\bibitem{borodin2012max}
A.~Borodin, H.~C. Lee, and Y.~Ye, ``Max-sum diversification, monotone
  submodular functions and dynamic updates,'' in \emph{Proceedings of the 31st
  ACM SIGMOD-SIGACT-SIGAI symposium on Principles of Database Systems}, 2012,
  pp. 155--166.

\end{thebibliography}
